\documentclass{article} %
\usepackage{iclr2027_conference,times}

\usepackage{amsmath,amsfonts,bm}

\def\eqref#1{equation~\ref{#1}}

\def\1{\bm{1}}

\DeclareMathAlphabet{\mathsfit}{\encodingdefault}{\sfdefault}{m}{sl}
\SetMathAlphabet{\mathsfit}{bold}{\encodingdefault}{\sfdefault}{bx}{n}

\usepackage{hyperref}
\usepackage{url}
\usepackage{booktabs}
\usepackage{graphicx}
\usepackage{capt-of}

\usepackage{listings}
\usepackage{titletoc}
\usepackage{enumitem}
\usepackage[table]{xcolor}
\definecolor{rowblue}{RGB}{229,241,251}

\definecolor{bestcell}{RGB}{225,239,248}

\usepackage[colorinlistoftodos]{todonotes}
\definecolor{todoblue}{RGB}{220,235,250}
\colorlet{punct}{red!60!black}
\definecolor{background}{HTML}{F8F8F8}
\definecolor{delim}{RGB}{20,105,176}
\colorlet{numb}{magenta!60!black}

\lstdefinelanguage{json}{
    basicstyle=\ttfamily\footnotesize,
    numbers=left,
    numberstyle=\scriptsize,
    stepnumber=1,
    numbersep=8pt,
    showstringspaces=false,
    breaklines=true,
    frame=single,
    backgroundcolor=\color{background},
    literate=
     *{0}{{{\color{numb}0}}}{1}
      {1}{{{\color{numb}1}}}{1}
      {2}{{{\color{numb}2}}}{1}
      {3}{{{\color{numb}3}}}{1}
      {4}{{{\color{numb}4}}}{1}
      {5}{{{\color{numb}5}}}{1}
      {6}{{{\color{numb}6}}}{1}
      {7}{{{\color{numb}7}}}{1}
      {8}{{{\color{numb}8}}}{1}
      {9}{{{\color{numb}9}}}{1}
      {:}{{{\color{punct}{:}}}}{1}
      {,}{{{\color{punct}{,}}}}{1}
      {\{}{{{\color{delim}{\{}}}}{1}
      {\}}{{{\color{delim}{\}}}}}{1}
      {[}{{{\color{delim}{[}}}}{1}
      {]}{{{\color{delim}{]}}}}{1},
}
\lstdefinestyle{prompt}{
    basicstyle=\ttfamily\scriptsize,
    breaklines=true,
    breakindent=0pt,
    columns=fullflexible,
    keepspaces=true,
    frame=single,
    backgroundcolor=\color{background},
    literate={—}{{---}}1,
}

\title{EPSBench: Benchmarking LLM Agents\\for Earnings Forecasting}

\author{Yanbin Yin\thanks{Equal contribution.} \quad Chuyu Xu\footnotemark[1] \quad Liqin Ye \quad Jing Peng \quad Chao Zhang \\
Georgia Institute of Technology \\
\texttt{\{yyin353,cxu433,liqiny,jpeng325,chaozhang\}@gatech.edu}
}

\newcommand{\aq}[1]{}

\iclrfinalcopy
\begin{document}

\maketitle

\begin{abstract}
Quarterly earnings forecasting requires an agent to connect financial evidence to the income-statement line items that produce a company's reported earnings. We introduce EPSBench, a benchmark for studying this ability across 149 firm-quarters from 86 US companies. Agents read a fixed collection of filings, earnings-call transcripts, and news, then estimate company-specific income-statement drivers with cited evidence. A shared accounting calculator converts these estimates into GAAP diluted earnings per share (EPS), providing a common basis for comparing forecast assumptions. We evaluate six models under a common harness, assessing point forecast error relative to date-matched analyst consensus and checking whether their prediction intervals achieve the stated coverage. Three models show no detectable paired difference in point forecast error from consensus, and three have higher error. Across all six models, nominal 80\% intervals contain reported EPS in only 37--70\% of runs that return an interval, and most misses fall above the upper bound because the forecasts run low. The driver records trace most of the error gap between models to operating-income estimates.
\end{abstract}

\section{Introduction}
\label{sec:intro}

Forecasts inform decisions that must be made before their outcomes are known, including investment and business planning \citep{ball_automated_2018,zou_forecasting_2022}. Some forecasting tasks ask only for a direction or a probability \citep{chen_predicting_2022,karger_forecastbench_2025}. Forecasting a reported quantity asks for a magnitude, which an agent must assemble from incomplete and sometimes conflicting evidence. An agent may find a company's sales guidance, recent cost increases, and historical financial statements, but must judge how much those facts move the next quarter's earnings, since higher sales need not imply higher profit if costs rise faster. A useful forecast puts a size on these effects and states what remains uncertain.

Recent benchmarks have made forecasting a substantial part of agent evaluation. ForecastBench elicits probabilities for unresolved events and compares models with human forecasters \citep{karger_forecastbench_2025}. FutureX evaluates search-enabled agents on diverse future events, including numerical financial predictions, while FinDeepForecast generates live corporate and macroeconomic forecasting tasks, including EPS \citep{zeng_futurex_2026,li_findeepforecast_2026}. For EPS, both ask the agent for the predicted figure itself, leaving the route to it to whatever the model writes in its trajectory. An analyst reaches EPS from below, estimating revenue, then the costs charged against it, and finally the items beneath them, with each estimate resting on a particular piece of evidence. A benchmark can ask an agent to forecast in the same order, so that those estimates and their evidence are recorded alongside the final number, which lets an error be traced to the stage of the statement where it enters.

\begin{figure}[!htb]
\centering
\includegraphics[width=0.98\textwidth]{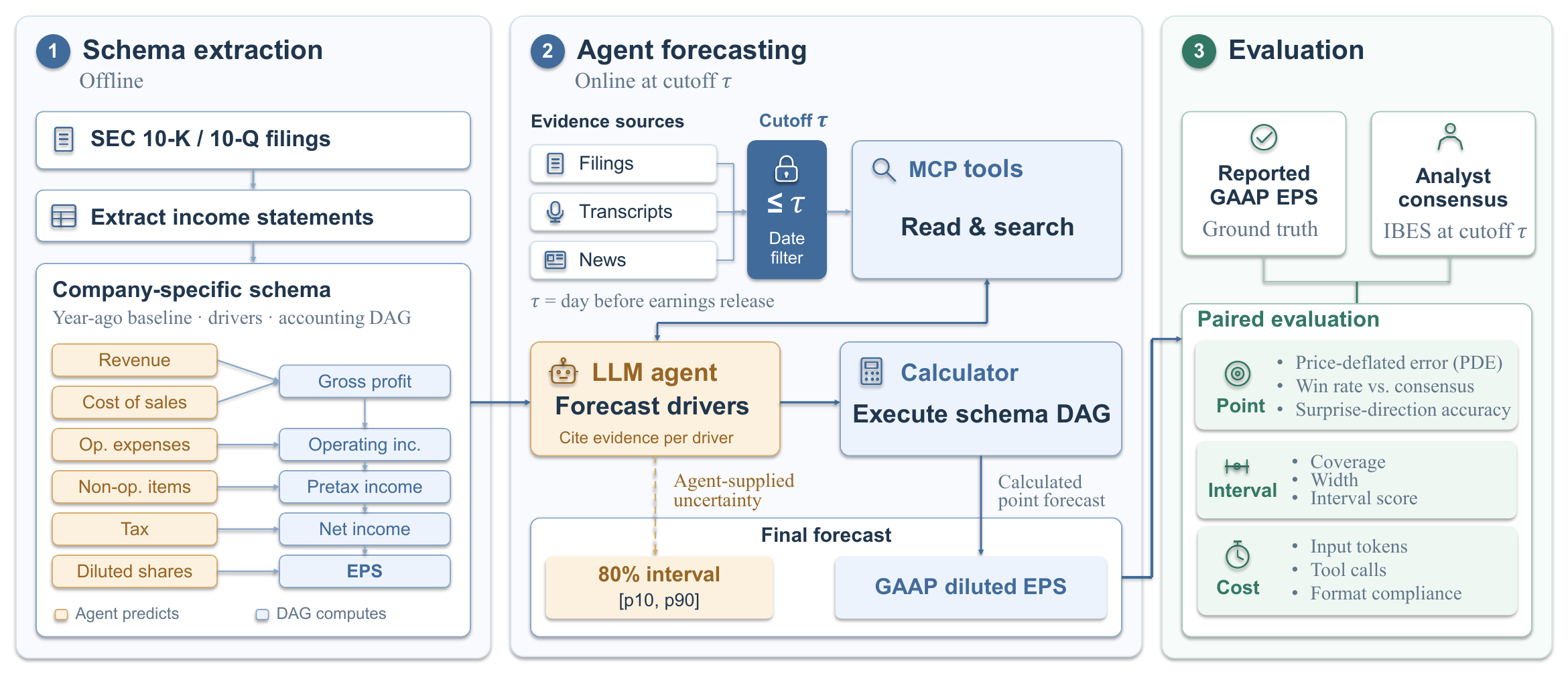}
\caption{\textbf{EPSBench workflow.} Left: historical filings are used to construct company-specific accounting schemas comprising a year-ago baseline, forecast drivers, and a directed acyclic graph of accounting identities. The DAG shown is illustrative; identities vary by firm-quarter. Center: agents use filings, earnings-call transcripts, and news available by the day before the earnings release to estimate drivers with cited evidence. A shared accounting calculator converts these estimates into GAAP diluted EPS, and agents provide an 80\% prediction interval. Right: forecasts are evaluated against reported EPS, using date-matched analyst consensus as a reference, alongside measures of interval calibration, resource use, and format compliance.}
\label{fig:pipeline}

\end{figure}

We introduce EPSBench, a benchmark that organizes quarterly earnings forecasting around each company's income statement. Earnings provide a recurring numerical target with a reported outcome, and the statement specifies how a company's reported line items combine into GAAP diluted earnings per share (EPS).\footnote{US companies prepare their filed financial statements under Generally Accepted Accounting Principles (GAAP). Companies and analysts also report adjusted or non-GAAP earnings that exclude selected items. EPSBench evaluates GAAP diluted EPS throughout, so agents must distinguish adjusted figures from the target when using them as evidence.} EPSBench contains 149 firm-quarters from 86 US companies. For each firm-quarter, we build a \emph{schema} from historical filings that identifies the quantities to forecast, which we call \emph{drivers}, and the accounting identities relating them. Agents work inside a fixed information environment built from filings, earnings-call transcripts, and news, all restricted to what was published before the results were released. They estimate these quantities, support their estimates with cited evidence, and submit them to a shared accounting calculator. The calculator returns the implied EPS, and the agent reports a prediction interval to express uncertainty about the outcome, as Figure~\ref{fig:pipeline} shows. We score the submitted EPS and its interval, and because the arithmetic is fixed, each score stays attached to the line-item estimates and cited evidence that produced it.

We evaluate six models from four vendors under a common harness that manages the model's tool calls and context, scoring forecasts against reported EPS and using date-matched analyst consensus as a professional reference. Three models, termed the leading models, show no detectable paired difference in point forecast error from consensus, and the other three have higher error. All six fall short of 80\% interval coverage, covering only 37--70\% of outcomes among runs that return an interval, and most misses fall above the upper bound consistent with low point forecasts, as consensus forecasts also do. The gap between the two groups opens on quarters whose earnings move far from the year-ago level, where the leading models more often adjust far enough and in the right direction, and it lies almost entirely in the operating lines of the income statement. Sections~\ref{sec:channels} and~\ref{sec:harness} compare information sources, harnesses, and reasoning effort: news and filings carry partly different information, while harness and effort change coverage more than point forecast error.

Our contributions are \textbf{(1)} company-specific accounting schemas that define the forecasting task and map driver estimates to EPS; \textbf{(2)} a common evaluation protocol linking dated evidence and driver estimates to point and interval scores; and \textbf{(3)} an empirical analysis that locates model differences in operating-income forecasts and assesses reported uncertainty. We release the schemas, tools, prompts, forecast records, scores, and trajectories to support reproducible comparisons.

\section{Related Work}
\label{sec:related}

\textbf{Forecasting with language model agents.}
Forecasting benchmarks increasingly evaluate how models use current information to predict unresolved outcomes. Retrieval-based systems combine dated evidence with probabilistic forecasts \citep{halawi_approaching_2024,zou_forecasting_2022,ye_alphabet_2026}; ForecastBench continuously elicits forecasts on unresolved questions, and Forecast-Dojo replays resolved prediction-market questions with dated news so that agents can forecast at successive historical dates \citep{karger_forecastbench_2025,ye_forecastdojo_2026}. FutureX, FinDeepForecast, and LiveMacroEval extend evaluation to search-enabled agents and to numerical financial and macroeconomic targets \citep{zeng_futurex_2026,li_findeepforecast_2026,zhao_livemacroeval_2026}. EPSBench focuses on a common representation of an earnings forecast: agents estimate the same company-specific drivers, whose accounting relationships determine EPS. The resulting records support comparisons of quantitative assumptions, while outcome-based metrics assess the point forecast and interval.

\textbf{Financial prediction and research.}
Financial benchmarks distinguish forecasting an unreported outcome from reasoning over information already disclosed. FinQA studies numerical reasoning over supplied reports, while FinAgentBench evaluates financial document retrieval \citep{chen_finqa_2021,choi_finagentbench_2025}. Prior earnings work predicts earnings direction or surprises and, in some cases, compares model forecasts with professional analysts \citep{ou_financial_1989,shu_fincall-surprise_2026,chen_predicting_2022,yin_complex_2026,an_simulating_2026}. FinanceHarness evaluates financial research reports under controlled model and harness configurations \citep{xiao_financeharness_2026}. EPSBench instead evaluates future earnings from point-in-time evidence through structured income-statement forecasts that determine GAAP diluted EPS.

\textbf{Uncertainty and evaluation validity.}
Forecast quality is not captured by point accuracy alone: useful forecasts must also represent uncertainty, motivating calibration measures and proper scoring rules \citep{gneiting_probabilistic_2007,gneiting_strictly_2007}. Agent evaluation further raises questions of execution cost, reliability, and intermediate progress \citep{kapoor_ai_2025,yao_taubench_2025,ma_agentboard_2024}. Retrospective forecasting introduces an additional validity challenge, because apparent performance can be inflated by training exposure, revised webpages, or misdated information \citep{paleka_pitfalls_2026,sarkar_lookahead_2024,lopez-lira_memorization_2025}. EPSBench addresses these concerns jointly by combining point and interval evaluation with execution metrics, tool-enforced point-in-time access, and a memorization test. Appendix~\ref{app:related} discusses earnings forecasting, financial research agents, and these evaluation issues in more detail.

\section{EPSBench}
\label{sec:design}

Each EPSBench instance asks an agent to forecast a company's GAAP diluted EPS for one quarter, using only documents available the day before the results are released. The agent reads filings, earnings-call transcripts, and news, submits a value for each driver in the company's schema with the evidence behind it, and reports an 80\% prediction interval for the EPS that the schema's accounting identities compute. We score the EPS and the interval against the outcome the company reported. Appendix~\ref{app:env} gives implementation details.

\subsection{Task Instances and Information Cutoff}
\label{sec:task}

Each instance pairs a company $c$ with a target fiscal quarter $q$, which we call a \emph{firm-quarter}. The target is GAAP diluted EPS, the measure reported in the company's filed income statement. We define the release date as $d^{\text{rel}}_{c,q}=\min(d^{\text{call}}_{c,q},d^{\text{ann}}_{c,q})$, taking the earlier of the earnings-call and official-announcement dates from the Institutional Brokers' Estimate System (IBES). This accounts for companies that publish their results before holding the call. The information cutoff and the date used for analyst consensus are
$$\tau=d^{\text{rel}}_{c,q}-1\text{ day}.$$
Because companies announce results weeks after a quarter closes, the target quarter has always ended by $\tau$: the task is to estimate a result that already exists but has not yet been published, as analysts do immediately before an announcement.

We select large, widely followed US companies across eight sectors for which historical filings, a prior-quarter transcript, and an IBES GAAP consensus are available. The benchmark contains 149 firm-quarters from 86 companies with announcements between 2 December 2025 and 24 June 2026, and this set is the common evaluation sample for every model, information-source, and harness comparison. Appendix~\ref{app:companies} provides the full list of companies, and Section~\ref{sec:setup} describes the separation between the announcements and the models' training cutoffs.

\subsection{Company-Specific Accounting Schemas}
\label{sec:offline_extraction}

Income statements differ across companies, so no single template of forecast quantities fits them all. A bank reports net interest income, for example, whereas a technology company may report separate product and service revenues. We therefore construct a schema for each firm-quarter from the company's own historical statements. The schema has three parts. The baseline holds the line-item values of the year-ago quarter. The drivers mostly state how a baseline line item changes, as a growth rate for a revenue line or a change in a cost ratio; the rest, such as other expense, tax, and share count, give an amount or rate directly. A directed acyclic graph (DAG) of accounting identities combines the resulting line items into diluted EPS, with the drivers setting its leaves. In Microsoft's fiscal 2026Q3 schema the drivers are product and service revenue growth, the change in cost of revenue as a share of total revenue, operating-expense growth, other expense, income tax expense, and diluted shares. Appendix~\ref{appendix:schema} gives the complete example.

We build each schema in several model stages. An LLM extractor first retrieves income-statement tables from the filings, and a reasoning model uses the four preceding quarterly statements to construct the DAG around the year-ago quarter, $q{-}4$. A further model extracts the historical tax and share-count references used as driver defaults. We then verify the schema by executing its DAG on the extracted year-ago values: 146 of 149 reproduce reported EPS within one cent, and the remaining three within four cents. Appendix~\ref{app:schema_construction} details the schema construction and validation.

\subsection{Agent Forecasting}
\label{sec:data}

An agent forecasts from a fixed information environment holding three kinds of source, each covering a different part of the evidence. Annual and quarterly filings (10-K and 10-Q) supply historical statements and detailed disclosures, the prior earnings call usually contains management's outlook for the target quarter, and news covers later developments while sometimes repeating the same guidance. We collect filings from SEC EDGAR, transcripts from Seeking Alpha, and news from CC-News, and expose them through six tools on a Model Context Protocol (MCP) server. The prompt directs agents to the $q{-}1$ call and the $q{-}1$ and $q{-}4$ filings, which provide the recent outlook and the statement underlying the schema; earlier documents remain available on request. Removing either news tools or filing and transcript tools creates the information conditions compared in Section~\ref{sec:channels}.

Every source stops at $\tau$. The server blocks the target-quarter filing and call, and news search covers $[\tau_0,\tau]$, where $\tau_0$ is the date of the $q{-}1$ call. The news tools apply the cutoff to the index shards and to each article they return, rejecting any article whose recorded effective date is later than $\tau$ or missing; Appendix~\ref{app:models_environment} describes how that date is determined and gives the retrieval configuration. Agents cannot read the stored scoring outcome, and the harness configurations in Section~\ref{sec:setup} block shell, file, and web access. We fix this collection because live search would return pages that postdate the announcement, and a released index lets others re-run the evaluation.

For each driver, the agent writes an \emph{evidence line} identifying its source and explaining the estimate, then submits the driver values to the calculator. The accounting calculator executes the schema and returns the income statement and EPS. The agent keeps that EPS as its point forecast and reports p10 and p90 bounds rather than moving the point to the interval midpoint. Each run is a single attempt without restarts; a final message that cannot be parsed into the required forecast counts as a format failure. Appendix~\ref{app:tools_protocol} provides details of the tools and run protocol.

Each run leaves a full trajectory: the agent's messages and tool calls, the documents each call returned, and the final driver values with their evidence lines. Because the calculator performs the arithmetic, the record also holds every intermediate node of the DAG.

\subsection{Evaluation Measures}
\label{sec:eval}

\textbf{Reported EPS and reference forecasts.}
We score every forecast against reported GAAP diluted EPS from the target-quarter income statement, verified against that filing. Three reference forecasts give the comparison a scale: year-ago EPS and last-quarter EPS measure simple historical extrapolation, and the analyst consensus, the IBES GAAP EPS consensus at $\tau$, provides a dated professional reference. The consensus comparison is not controlled for information access: analysts have access to sources outside the benchmark, while agents may encounter analyst expectations in retrieved news even though the tools do not return the IBES series. The prompt allows the agent to use an encountered consensus figure only as a sanity check after it has set its drivers. This mirrors the analyst's own position, since analysts also see one another's estimates; EPSBench instead requires every driver to be supported by cited evidence.

\textbf{Point forecast.}
Our primary measure is price-deflated error, $\mathrm{PDE}=|\hat y-y|/P$, where $\hat y$ is the forecast EPS, $y$ is reported EPS, and $P$ is the closing share price on the last trading day before release. Scaling by price avoids unstable percentage errors when EPS is close to zero and follows common practice in analyst-forecast evaluation \citep{hong_analyzing_2003}. The win rate is the fraction of instances on which a forecast has lower PDE than consensus, with an unparseable run counted as a loss. Surprise-direction accuracy is the fraction of instances on which the forecast correctly predicts whether reported EPS lands above or below consensus.

\textbf{Interval forecast.}
We assess the reported p10--p90 interval by its coverage, the fraction of outcomes it contains, together with the two tail rates, its width, and its interval score. The score adds the width to a penalty for outcomes outside the interval; it is proper, so misreporting p10 or p90 cannot improve the expected score \citep{gneiting_strictly_2007}. Like PDE, it is scaled by share price, as defined in Appendix~\ref{app:evaluation_details}. We elicit the 10th and 90th percentiles, the P10--P90 range commonly used to report forecast uncertainty; with 447 runs per model it leaves about 45 expected misses on each side, enough to estimate the two tail rates separately.%

\textbf{Aggregation and uncertainty.}
We run each configuration three times per instance. For point forecasts, we take the median PDE over each instance's scoreable runs and report the median across the 149 instances. For interval forecasts, we pool all runs that return an interval across instances, reporting coverage and tail rates as shares of those runs, the median width, and the mean interval score \citep{gneiting_strictly_2007}; Table~\ref{tab:calib} in Appendix~\ref{app:results} reports these run counts by model. To compare two forecasters, we pair them on each instance, take the difference in their PDEs, and report the median of the 149 differences with a bootstrap 90\% confidence interval. Pairing removes the variation in difficulty across instances, which is much larger than the differences between forecasters. The bootstrap resamples instances with replacement 10,000 times and takes the 5th and 95th percentiles of the medians. An interval containing zero means that the data do not resolve the difference at this sample size; it does not establish equivalence. We also record model requests, tool calls, input, output, and reasoning tokens, and format compliance; Appendix~\ref{app:evaluation_details} gives their aggregation rules.

\section{Experimental Setup}
\label{sec:setup}

The main experiment compares six models from four vendors, each at its provider's default reasoning effort, on the same 149 tasks, accounting schemas, information environment, and evaluation protocol. Table~\ref{tab:models} in Appendix~\ref{app:models} lists their release dates and published knowledge cutoffs, and we abbreviate them as Opus 4.6, GPT-5.5, GPT-5.4, GPT-5-mini, Qwen 3.5, and GLM 4.5 Air. Additional experiments hold the model fixed while changing the information available to GPT-5.5 or the harness and reasoning effort used by GPT-5.5 and Opus 4.6. We ran all configurations three times per firm-quarter, yielding 447 runs per configuration.

\textbf{Harnesses and model settings.}
The primary harness is OpenCode SDK 1.17.9 with shell, file, and web tools disabled, leaving the six MCP tools as the agent's interface. We also test a vanilla harness built on the OpenAI SDK for GPT-5.5, and Claude Code 2.1.269 for Opus 4.6. Section~\ref{sec:harness} additionally runs Opus 4.6 under both harnesses and GPT-5.5 under OpenCode at maximum effort. Appendix~\ref{app:harness} details the harness configurations and isolation controls. All harnesses receive the same prompts, provided in Appendix~\ref{app:prompts}, apart from the stated changes required when an information source is removed. Source ablations remove news, filings and transcripts, or both, and update the prompts to name the remaining tools.

\textbf{Temporal separation from training data.}
All benchmark announcements follow the published knowledge cutoffs of four models. Of the two that do not disclose a cutoff, GLM 4.5 Air was released before the evaluation window, whereas Qwen 3.5 was released within it. We additionally conduct a memorization test on Qwen 3.5, finding no detectable recall of benchmark outcomes under the tested protocol. Appendix~\ref{app:probe} provides the test details and results.

\section{Results}
\label{sec:results}

\subsection{Point Forecasts}
\label{sec:point}

\begin{figure}[!htb]
\centering
\includegraphics[width=0.65\textwidth]{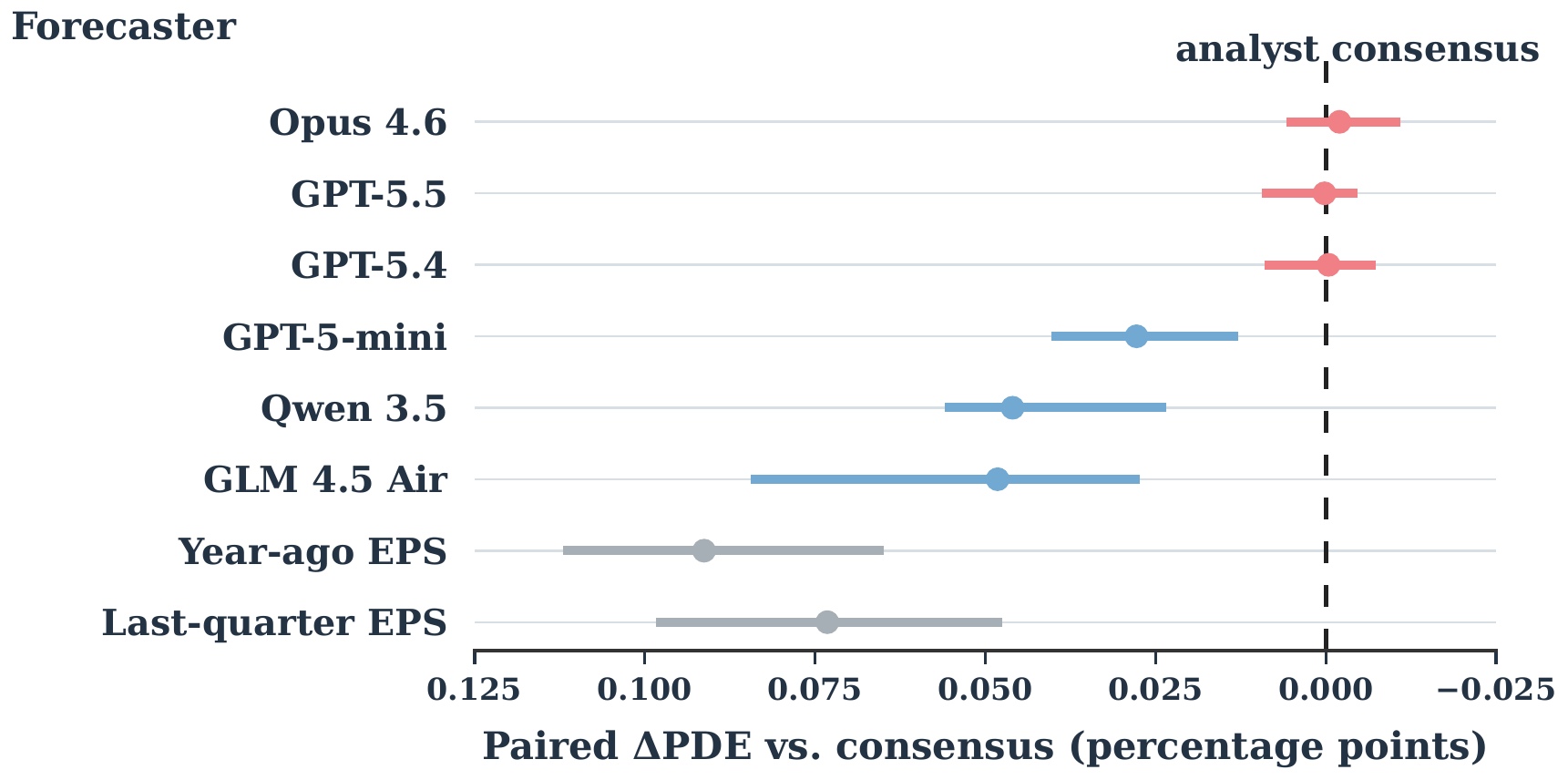}
\caption{\textbf{Paired point forecast error comparisons with analyst consensus.} Dots show the median paired difference in PDE between each forecaster and analyst consensus on 149 firm-quarters; horizontal lines show 90\% bootstrap confidence intervals. Gray rows are the two historical reference forecasts. Negative values, to the right of the dashed line, indicate lower error than consensus.}
\label{fig:forest}
\end{figure}

Against analyst consensus, the six models fall into two groups, which we call the leading models (Opus 4.6, GPT-5.5, and GPT-5.4) and the trailing models (GPT-5-mini, Qwen 3.5, and GLM 4.5 Air). Table~\ref{tab:main} shows that Opus 4.6, GPT-5.5, and GPT-5.4 have median PDEs of 0.113--0.131\%, close to the 0.111\% of consensus, and each beats consensus on 50\% of instances; the other three reach 0.190--0.215\% and win on 28--38\%. Surprise-direction accuracy follows the same split, from 63\% for GPT-5.5 to 45\% for GLM 4.5 Air. All six improve on copying year-ago or last-quarter EPS (0.238\% and 0.226\%). Figure~\ref{fig:forest} tests these differences with paired comparisons. The intervals for Opus 4.6, GPT-5.5, and GPT-5.4 all include zero, and Opus 4.6 has the lowest median paired difference, $-0.002$ percentage points (pp) of PDE with a 90\% interval of $[-0.011,+0.006]$. The other three have higher paired error than consensus.

\begin{table}[!htb]
\caption{\textbf{Main results.} Six models and three reference forecasters on 149 firm-quarters under OpenCode with the full information environment. PDE is the median price-deflated error; Win is the share of instances with lower PDE than analyst consensus; Surprise is surprise-direction accuracy. Interval metrics pool runs returning a p10--p90 interval; width and score are scaled by share price, in percentage points (pp). Shading marks the leading models; bold marks each column's best model value, with ties, except width, which has no single best value. Section~\ref{sec:eval} details the aggregation.}
\label{tab:main}
\begin{center}
\small
\setlength{\tabcolsep}{8pt}
\begin{tabular}{@{}l rrr rrr@{}}
\toprule
 & \multicolumn{3}{c}{Point forecast} & \multicolumn{3}{c}{80\% interval forecast} \\
\cmidrule(lr){2-4}\cmidrule(l){5-7}
Forecaster & \shortstack{PDE\\(\%) $\downarrow$} & \shortstack{Win\\(\%) $\uparrow$} & \shortstack{Surprise\\(\%) $\uparrow$} & \shortstack{Coverage\\(\%)} & \shortstack{Median \\Width (pp)} & \shortstack{Mean interval \\Score (pp) $\downarrow$} \\
\midrule
Analyst consensus & 0.111 & -- & -- & -- & -- & -- \\
Year-ago EPS & 0.238 & 21 & 36 & -- & -- & -- \\
Last-quarter EPS & 0.226 & 27 & 51 & -- & -- & -- \\
\midrule
\rowcolor{bestcell}\multicolumn{1}{@{}>{\columncolor{bestcell}[0pt][\tabcolsep]}l}{Opus 4.6} & \textbf{0.113} & \textbf{50} & 62 & \textbf{70} & 0.41 & \multicolumn{1}{>{\columncolor{bestcell}[\tabcolsep][0pt]}r@{}}{\textbf{2.87}} \\
\rowcolor{bestcell}\multicolumn{1}{@{}>{\columncolor{bestcell}[0pt][\tabcolsep]}l}{GPT-5.5} & 0.131 & \textbf{50} & \textbf{63} & 61 & 0.33 & \multicolumn{1}{>{\columncolor{bestcell}[\tabcolsep][0pt]}r@{}}{3.08} \\
\rowcolor{bestcell}\multicolumn{1}{@{}>{\columncolor{bestcell}[0pt][\tabcolsep]}l}{GPT-5.4} & 0.126 & \textbf{50} & 60 & 52 & 0.25 & \multicolumn{1}{>{\columncolor{bestcell}[\tabcolsep][0pt]}r@{}}{4.29} \\
GPT-5-mini & 0.191 & 38 & 51 & 50 & 0.39 & 19.61 \\
Qwen 3.5 & 0.190 & 32 & 47 & 37 & 0.26 & 4.87 \\
GLM 4.5 Air & 0.215 & 28 & 45 & 44 & 0.33 & 6.48 \\
\bottomrule
\end{tabular}
\end{center}
\end{table}

The pairwise comparisons in Table~\ref{tab:pairwise} support this separation. Each leading model has lower point forecast error than each trailing model, while no two leading models differ significantly from each other. The trailing models also have more dispersed errors: their interquartile ranges are 0.34--0.44 pp, against 0.23--0.26 pp for the leading models, and Table~\ref{tab:eff} shows that each trailing model has a PDE above 1\% on 13--23 instances, compared with 8--11 instances for each leading model.

\subsection{Interval Forecasts}
\label{sec:calibration}

\begin{figure}[!htb]
\centering
\includegraphics[width=0.64\textwidth]{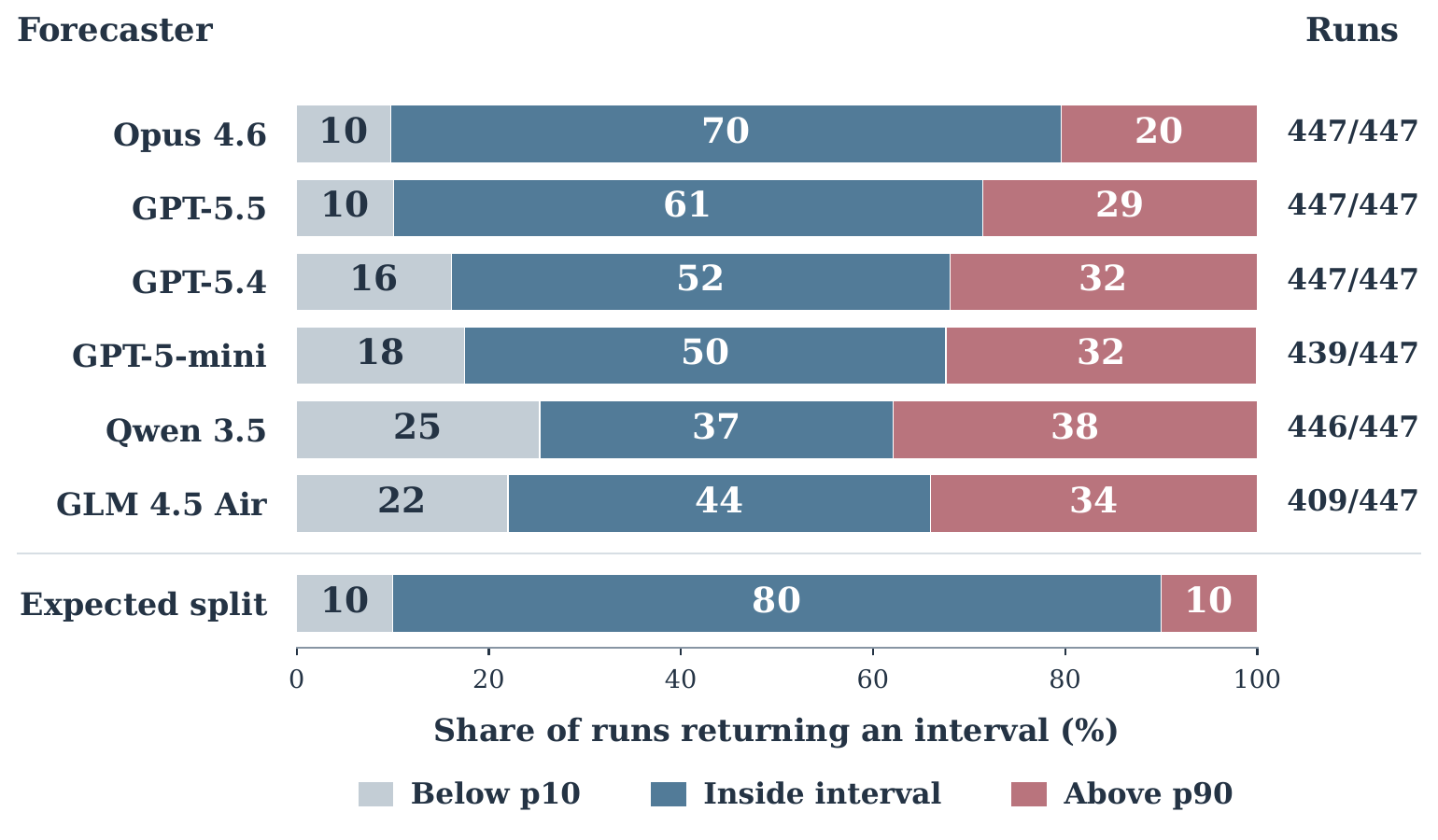}
\caption{\textbf{Prediction interval coverage and tail rates.} Stacked bars show the percentages of reported EPS outcomes below, inside, and above each model's p10--p90 prediction interval, pooling runs that return an interval. The Runs column lists the number of returned intervals over total runs. The expected split is 10\% below, 80\% inside, and 10\% above the interval.}
\label{fig:calib}
\end{figure}

An 80\% prediction interval should contain reported EPS in 80\% of outcomes, with 10\% falling on each side. Table~\ref{tab:main} shows that no model reaches this coverage: it ranges from 37\% for Qwen 3.5 to 70\% for Opus 4.6. Figure~\ref{fig:calib} shows that the misses are lopsided. For every model, reported EPS lands above the upper bound in 20--38\% of outcomes, well over the nominal 10\%, while below the lower bound Opus 4.6 and GPT-5.5 stay at 10\% and the other four reach 16--25\%.

The misses lean upward because the point forecasts themselves tend to be too low. Table~\ref{tab:calib} in Appendix~\ref{app:results} reports the median signed error $(y-\hat y)/P$, which is positive when a forecast falls below reported EPS: it ranges from $+0.021$ to $+0.048$ pp across models, and its confidence interval excludes zero for every model. Each interval is nearly symmetric around its point forecast, with midpoint displacements within 0.004 pp, so a low forecast pulls the whole interval down with it. Consensus is also too low, at $+0.028$ pp with a confidence interval of $[+0.008,+0.051]$. Prior work documents the walk-down of analyst forecasts to beatable targets \citep{richardson_walk-down_2004} and the rewards to meeting or beating expectations and their rise over time \citep{bartov_rewards_2002,brown_temporal_2001}. Expectation management and exposure to analyst expectations could both produce the agents' bias.

Table~\ref{tab:main} also shows that coverage alone does not rank the intervals. GPT-5.4 and GPT-5.5 have similar point forecast errors, but GPT-5.4's intervals are narrower (median width 0.25 against 0.33 pp), cover fewer outcomes (52\% against 61\%), and receive a worse mean interval score (4.29 against 3.08). GPT-5-mini shows the opposite case: its coverage of 50\% is close to GPT-5.4's, yet its mean score of 19.61 is three to seven times that of every other model (2.87--6.48). A few of its intervals are narrow and far from reported EPS, and because the score charges ten times the distance of each miss, those runs dominate the mean.

\subsection{Analysis of Forecast Differences}
\label{sec:departure}

Model differences are largest on quarters where earnings move far from their year-ago level. For each instance, we measure how much reported EPS changed from the year-ago quarter, scaled by share price, and compare the third of instances with the smallest changes (near) against the third with the largest (far). On near quarters last year's EPS is already a good forecast, so the question is whether a forecaster can anticipate a large change. Table~\ref{tab:departure} shows that on the 49 near instances, model median PDEs range from 0.048\% to 0.082\%, compared with 0.036\% for consensus. On the 50 far instances, Opus 4.6, GPT-5.5, and GPT-5.4 reach 0.392--0.398\%, compared with 0.599--0.716\% for the other three models and 0.409\% for consensus.

\begin{table}[!htb]
\caption{\textbf{Forecast performance by departure from year-ago EPS.} Near and far groups are the thirds of instances with the smallest and largest price-scaled EPS changes from the year-ago quarter, defined from reported outcomes; the year-ago EPS row gives the median change. Median PDE is aggregated as in Section~\ref{sec:eval}. Numeric evidence is the share of far-instance runs that cite a number for their most-adjusted driver. Error halved counts far instances whose median error across the three runs is at most half that of year-ago EPS, and Wrong direction counts far instances where at least two of the three runs move away from year-ago EPS opposite to the realized change.}
\label{tab:departure}
\begin{center}
\small
\setlength{\tabcolsep}{5pt}
\begin{tabular}{@{}l rr rrr@{}}
\toprule
 & \multicolumn{2}{c}{Median PDE (\%)} & \multicolumn{3}{c}{Far instances only ($n=50$)} \\
\cmidrule(lr){2-3}\cmidrule(l){4-6}
Forecaster & \shortstack{Near\\($n=49$)} & \shortstack{Far\\($n=50$)} & \shortstack{Numeric\\evidence (\%)} & \shortstack{Error\\halved ($n$)} & \shortstack{Wrong\\direction ($n$)} \\
\midrule
Year-ago EPS & 0.080 & 0.975 & -- & -- & -- \\
Analyst consensus & 0.036 & 0.409 & -- & -- & -- \\
\midrule
Opus 4.6 & 0.048 & 0.396 & 95 & 30 & 5 \\
GPT-5.5 & 0.053 & 0.392 & 96 & 30 & 6 \\
GPT-5.4 & 0.050 & 0.398 & 95 & 29 & 5 \\
GPT-5-mini & 0.082 & 0.614 & 62 & 19 & 12 \\
Qwen 3.5 & 0.071 & 0.599 & 72 & 17 & 12 \\
GLM 4.5 Air & 0.074 & 0.716 & 35 & 12 & 12 \\
\bottomrule
\end{tabular}
\end{center}
\end{table}

Table~\ref{tab:departure} also shows how the two groups handle the far instances. The leading models more often move their forecasts far enough from last year's EPS to cut its error at least in half, on 29--30 of the 50 instances against 12--19 for the trailing models, and they move in the wrong direction less often, on 5--6 instances against 12. To see what supports these adjustments, we read the evidence line for the driver that each run adjusts most, since a large departure from last year needs a concrete basis such as management guidance or a disclosed transaction amount. The leading models cite a number there in 95--96\% of runs, against 35--72\% for the trailing models; Appendix~\ref{app:results} gives the coding rules.

The records also show where in the income statement the two groups' errors
differ. On the 125 instances with an operating-income subtotal, the gap lies
almost entirely in operating income: median operating error is 0.089--0.098 pp
for the leading models versus 0.149--0.166 pp for the trailing models, while
the non-operating increments are similar at 0.007--0.020 pp.
Appendix~\ref{app:results} reports the full decomposition in
Table~\ref{tab:layers}. The records can further trace these errors to individual
drivers: in Boeing's fourth quarter of 2025, much of the gap comes from the gain
on a divestiture that Boeing reports within operating income.
Appendix~\ref{app:boeing_case} gives this case in detail.

\subsection{Contributions of Filings, Transcripts, and News}
\label{sec:channels}

We test how much GPT-5.5's error depends on news and on filings and transcripts. Table~\ref{tab:info_ablation} shows that, with the schema and accounting calculator held fixed, we detect no paired difference when only news is kept: the median change is $-0.000$ pp with a 90\% interval of $[-0.010,+0.009]$, while input use falls to about one quarter. Removing news and keeping filings and transcripts makes forecasts slightly worse, raising error by $+0.009$ pp $[+0.002,+0.016]$. Removing both makes them substantially worse: error rises by $+0.033$ pp $[+0.019,+0.047]$ and the win rate against consensus falls from 50\% to 31\%.

\begin{table}[!htb]
\caption{\textbf{Information-source ablation} for GPT-5.5 under OpenCode on 149 firm-quarters. Full includes filings, earnings-call transcripts, and news; PDE and Win are as in Table~\ref{tab:main}. Paired $\Delta$PDE is relative to Full, and positive values indicate higher error. Tools and Input are per-run medians, with Input including cached tokens.}
\label{tab:info_ablation}
\begin{center}
\footnotesize
\setlength{\tabcolsep}{4pt}
\begin{tabular}{@{}l rr r rr@{}}
\toprule
Information & \shortstack{PDE (\%) $\downarrow$} & \shortstack{Win (\%) $\uparrow$} & \shortstack{Paired $\Delta$PDE, pp vs.\ full (90\% CI)} & Tools & \shortstack{Input (K)} \\
\midrule
Full & 0.131 & 50 & -- & 47 & 837 \\
News only & 0.141 & 40 & $-0.000$ $[-0.010,+0.009]$ & 28 & 196 \\
Filings + transcripts & 0.138 & 42 & $+0.009$ $[+0.002,+0.016]$ & 25 & 376 \\
Neither & 0.210 & 31 & $+0.033$ $[+0.019,+0.047]$ & 1 & 8 \\
\bottomrule
\end{tabular}
\end{center}
\end{table}

To see which instances drive this result, we count an error increase as harmful when it exceeds 0.057 pp, the median spread of PDE across an instance's three runs. Table~\ref{tab:source_groups} in Appendix~\ref{app:results} sorts the 149 instances into three groups. On 67, no removal is harmful. On 32, neither single removal is harmful but removing both is. The remaining 50 depend on a particular source: 17 are hurt only by removing filings and transcripts, 24 only by removing news, and 9 by either removal. Dependence on the two sources is uncorrelated: their error increases have a rank correlation of $-0.01$.

On the 32 instances where only removing both is harmful, the sources carry overlapping information: management's guidance appears in the earnings call and reappears in news and earnings previews, and in inspected trajectories agents choose similar driver values while citing the transcript in one condition and an article in another. The 50 instances that depend on a particular source show that each source also carries information the other lacks: news supplies post-filing developments and current market prices, whereas filings supply detailed accounting quantities such as the book value of a divested business or the magnitude of an equity-method loss. Removing either source alone therefore barely moves the median error, yet for a third of the instances a particular source cannot be replaced.

\begin{table}[!htb]
\caption{\textbf{Harness and reasoning-effort comparisons} on 149 firm-quarters. Paired $\Delta$PDE is relative to each model's first row, OpenCode at the default effort, and coverage pools runs returning an interval. Tools, Input, and Output are per-run medians; Input includes cached tokens and Output includes reasoning tokens.}
\label{tab:harness}
\begin{center}
\footnotesize
\setlength{\tabcolsep}{3pt}
\begin{tabular}{@{}lllrrrrrr@{}}
\toprule
Model & Harness & Effort & \shortstack{PDE\\(\%)$\downarrow$}
& \shortstack{Paired $\Delta$PDE, pp\\(90\% CI)}
& \shortstack{Coverage\\(\%)} & Tools
& \shortstack{Input\\(K)} & \shortstack{Output\\(K)} \\
\midrule
Opus 4.6 & OpenCode & default & 0.113 & -- & 70 & 35 & 1056 & 9.7 \\
& Claude Code & default & 0.117 & $+0.001$ $[-0.005,+0.004]$ & 65 & 34 & 1029 & 36.4 \\
& OpenCode & max & 0.106 & $-0.002$ $[-0.006,+0.001]$ & 65 & 34 & 1011 & 40.5 \\
& Claude Code & max & 0.114 & $-0.004$ $[-0.008,+0.002]$ & 65 & 34 & 1032 & 40.4 \\
\midrule
GPT-5.5 & OpenCode & default & 0.131 & -- & 61 & 47 & 837 & 6.2 \\
& Vanilla & default & 0.110 & $-0.003$ $[-0.007,+0.002]$ & 54 & 38 & 561 & 8.6 \\
& OpenCode & max & 0.135 & $-0.000$ $[-0.003,+0.003]$ & 54 & 48 & 821 & 12.6 \\
\bottomrule
\end{tabular}
\end{center}
\end{table}

\subsection{Sensitivity to the Agent Harness}
\label{sec:harness}

Finally, we ask whether these results depend on the agent harness and reasoning effort. Table~\ref{tab:harness} shows that changing either barely moves point forecast error: relative to OpenCode at the default effort, median paired PDE changes by $-0.004$ to $+0.001$ pp, and all five intervals include zero. Resource use, by contrast, changes considerably. Maximum effort raises median output tokens per run, including reasoning, from 9.7K to 40.5K for Opus 4.6 and from 6.2K to 12.6K for GPT-5.5 under OpenCode. Claude Code already produces 36.4K at the default effort, close to its maximum-effort level, and for GPT-5.5 the vanilla harness uses 33\% fewer input tokens. Interval coverage also varies despite similar point forecast error: for Opus 4.6, it is 70\% under OpenCode at the default effort and 65\% in the other three configurations, while for GPT-5.5 it ranges from 54\% to 61\%.

\section{Conclusion}
\label{sec:conclusion}

EPSBench provides an accounting task for studying how agents turn financial evidence into earnings forecasts. The leading agents show no detectable difference in point forecast error from date-matched analyst consensus, yet all six produce intervals with coverage below 80\%; more outcomes fall above the upper bound than below the lower bound. Model differences are largest when earnings move sharply from the year-ago level, with most of the gap concentrated at the operating-income stage. Information access affects point forecast error and resource use, while harness and reasoning effort change resource use and interval coverage more than point forecast error.

\subsection*{AI use statement}

We used generative AI in schema extraction, benchmark implementation, and manuscript drafting and revision. The evaluated systems are themselves language model agents, and all reported experimental results come from recorded runs and evaluation scripts. The authors retain responsibility for validating the implementation and ensuring that the manuscript accurately represents the underlying results.

\subsection*{Ethics statement}

This study evaluates company financial information and model outputs without recruiting human participants. We obtain public filings from the SEC, licensed earnings-call transcripts from Seeking Alpha, news from CC-News, analyst consensus through an academic IBES/WRDS license, and share prices from Yahoo Finance. We use the IBES records for evaluation and report the forecasts as research outputs; this benchmark does not establish their suitability for investment decisions. The authors declare no conflict of interest.

\subsection*{Reproducibility statement}

Sections~\ref{sec:task}--\ref{sec:eval} define the task and evaluation protocol, Section~\ref{sec:setup} specifies the model and harness configurations, and Appendix~\ref{app:probe} describes the recall probes. We release the extracted schemas, allowing others to reproduce the evaluation without regenerating them. The original schema-generation model, \texttt{deepseek-reasoner}, which we used in June and August 2026, is no longer served; extending the benchmark to new quarters will require another extractor.

The release, available at \url{https://github.com/JadeXu16/epsbench}, includes the tool server, harness adapters, prompts, evaluation scripts, share prices, per-instance forecasts, reported EPS, and scores. We provide trajectories with transcript text replaced by document identifiers to respect licensing restrictions, along with fetch scripts for public filings and the filtered CC-News index. We do not release IBES consensus values: we report aggregate comparisons and, for each instance, whether the agent outperformed consensus.

\bibliography{iclr2027_conference,extra}
\bibliographystyle{iclr2027_conference}

\clearpage
\appendix
\renewcommand{\thesection}{\Alph{section}}

\begin{center}
    {\Large \textbf{Appendix for EPSBench}}
\end{center}

\startcontents[sections]
\printcontents[sections]{l}{1}{\setcounter{tocdepth}{2}}

\section{Additional Related Work}
\label{app:related}

\textbf{Earnings forecasting.}
Comparisons between automated earnings forecasts and analyst forecasts depend on what information the methods receive and what they are asked to predict. Early models used adjacent-quarter and seasonal patterns, while analysts benefited in part from information that arrived before they issued their forecasts \citep{foster_quarterly_1977,brown_univariate_1979,brown_security_1987}. Subsequent work used financial-statement information to predict the direction of earnings changes and construct stock trading strategies \citep{ou_financial_1989}. Comparisons with analyst forecasts show that relative performance depends on the forecasting method and evaluation setting \citep{chen_predicting_2022,ball_automated_2018,elend_earnings_2020}. LLM-based studies extend these approaches to direction and earnings-surprise prediction from statements and calls, and to direct EPS prediction from structured inputs that include analyst forecasts \citep{yin_complex_2026,shu_fincall-surprise_2026,an_simulating_2026}. Search-enabled forecasting systems further allow models to gather evidence when predicting financial quantities \citep{zeng_futurex_2026,li_findeepforecast_2026}. EPSBench follows this numerical forecasting tradition while specifying a common accounting task: agents estimate company-specific income-statement drivers from retrieved documents, and we evaluate the resulting quarterly GAAP EPS forecasts and intervals.

\textbf{Financial research and agent environments.}
Interactive benchmarks extend evaluation from reasoning over supplied information to acquiring that information and acting on it through tools. FinQA studies numerical reasoning over provided financial reports, whereas FinAgentBench evaluates retrieval of relevant financial documents and passages \citep{chen_finqa_2021,choi_finagentbench_2025}. More general environments such as AgentBench and WebArena assess multi-step interaction, and studies of agent interfaces and scaffolds show how the surrounding harness can affect performance \citep{liu_agentbench_2024,zhou_webarena_2024,yang_swe-agent_2024,zhang_cybench_2025,kapoor_holistic_2026}. FinanceHarness brings controlled model and harness comparisons to financial research reports using point-in-time data and expert-validated rubrics \citep{xiao_financeharness_2026}. EPSBench applies a related experimental approach to a numerical outcome: the main comparison holds the task, tools, accounting calculator, and harness fixed, while additional comparisons change the information sources or harness for selected models.

\textbf{Temporal validity of forecasting evaluations.}
A forecasting evaluation must distinguish information available at the forecast date from information that becomes available afterward. ForecastBench collects predictions before questions resolve, and LiveMacroEval records nowcasts in real time before each official release, while retrieval-based forecasting systems and replayable environments such as Forecast-Dojo restrict evidence by publication date \citep{karger_forecastbench_2025,zhao_livemacroeval_2026,halawi_approaching_2024,zou_forecasting_2022,ye_forecastdojo_2026}. Retrospective studies face additional difficulties because pretrained models may have encountered the outcomes and web pages may have been updated after their recorded dates \citep{paleka_pitfalls_2026,sarkar_lookahead_2024,lopez-lira_memorization_2025}. Benchmarks built from recent sources and studies using news after model cutoffs reduce some of these risks, but recency alone does not establish that a particular evaluation is free of exposure \citep{white_livebench_2025,lopez-lira_can_2026}. EPSBench uses document-date restrictions and examines the separation between model training and outcome announcements.

\textbf{Uncertainty and interpretation of agent forecasts.}
Forecast evaluation and agent evaluation provide complementary criteria for assessing uncertainty, resource use, and the steps leading to an estimate. In probabilistic forecasting, calibration and sharpness jointly characterize uncertainty, and proper scoring rules provide incentives for truthful quantile reports \citep{gneiting_probabilistic_2007,gneiting_strictly_2007}. Studies of verbalized LLM confidence document overconfidence, while agent evaluations also measure execution cost and reliability across runs \citep{xiong_can_2024,kapoor_ai_2025,yao_taubench_2025}. AgentBoard and ReCEval examine intermediate progress and reasoning steps, while structured representations connect numerical answers to recorded computations \citep{ma_agentboard_2024,prasad_receval_2023,lyu_faithful_2023}. EPSBench brings these forms of assessment to the same forecasting task: interval metrics evaluate stated uncertainty, and driver records support descriptive comparisons of the assumptions behind EPS estimates. The accounting calculator fixes the relationship between submitted drivers and EPS; it does not establish that the recorded explanations faithfully describe the model's internal reasoning.

\section{Environment and Protocol Details}
\label{app:env}

We provide the implementation details for the environment and evaluation protocol introduced in Section~\ref{sec:design}. We first describe how we build the forecasting schemas and prepare the documents, then specify the tools, run protocol, and evaluation measures.

\subsection{Schema Construction and Validation}
\label{app:schema_construction}

\textbf{Schema extraction.} We use o4-mini to extract income-statement tables from each 10-K and 10-Q. DeepSeek's \texttt{deepseek-reasoner} then reads the four quarterly statements preceding the target, $q{-}1$ through $q{-}4$, and generates the baseline leaves, formula DAG, and drivers described in Section~\ref{sec:offline_extraction}. When $q{-}4$ is a fiscal fourth quarter without a standalone statement, we subtract the first nine months from the annual figures in the 10-K. We obtain those nine-month figures from the cumulative column of the third-quarter 10-Q when available; otherwise, we sum the first three quarterly statements. We use GPT-5.4 to extract the tax and share-count references.

\textbf{Driver forms.} Let $V_q$ be a line item in the target quarter and $V_{q-4}$ its value in the year-ago baseline quarter. Let $A_q$ and $A_{q-4}$ denote the corresponding values of an anchor line item, typically revenue. The agent predicts a growth rate $g$, a margin change $\delta$ in basis points, an absolute value, or an effective tax rate, and each driver takes one of four forms:

\begin{itemize}
    \item \textbf{Year-over-year growth} (\texttt{yoy}), for scale line items such as segment revenues and R\&D: $V_q = V_{q-4}\,(1 + g)$, with $g$ a decimal fraction.
    \item \textbf{Margin change in basis points} (\texttt{bps}), for cost items that scale with the anchor: $V_q = A_q\,\big(V_{q-4}/A_{q-4} + \delta/10{,}000\big)$, with $\delta$ an integer number of basis points, so that $\delta = -20$ lowers the cost ratio by 0.20 percentage points.
    \item \textbf{Absolute value} (\texttt{absolute}), for non-operating and capital-structure items such as interest, other income, income tax expense, the diluted share count, non-controlling interest, and preferred dividends.
    \item \textbf{Rate} (\texttt{rate\_override}), for an effective tax rate applied to pretax income.
\end{itemize}

Dividing by 10,000 converts the integer basis-point inputs requested by the prompts into decimal changes in margins, as the prompts in Appendix~\ref{app:prompts} specify.

\textbf{Tax and share-count references.} To convert pretax income into diluted EPS, the agent must also forecast tax expense and the diluted share count. Unlike revenue and cost lines, neither follows from how the business is performing: the tax rate depends on tax law and one-off tax events, and the share count on buybacks and issuance, so the documents that describe the business say little about either. GAAP tax rates can vary sharply when a company records a one-off benefit or earns little pretax income. Share counts usually change more gradually through buybacks and issuance, making recent values useful references. The schema represents tax as an effective rate in 138 of the 149 instances, as an absolute expense in 10, and as both in one. The Microsoft example in Appendix~\ref{appendix:schema} uses the absolute form.

Three historical references accompany the share-count driver in every schema, the tax-rate driver in all 139 schemas that use one, and the absolute tax driver in 4 of the 11 that use one: the most recent quarter (\texttt{last\_q}), the trailing twelve months (\texttt{ttm}), and the year-ago quarter (\texttt{q4}), taken from the company's GAAP statements. We use their median as the default to reduce sensitivity to a single unusual quarter, and allow the agent to select a reference or provide its own value. The other seven absolute tax drivers default to the year-ago expense. We exclude adjusted tax-rate guidance from these references because it omits items that enter the GAAP earnings we evaluate.

\textbf{Schema verification.} We execute each schema's DAG on its extracted $q{-}4$ leaf values and compare the result with the EPS reported for that quarter. Of the 149 instances, 146 reproduce EPS within one cent and the other three within four cents. Rounding individual line items can accumulate in the final EPS calculation. We release the schemas together with the per-instance verification results, so that the evaluation can be reproduced without regenerating them.

\subsection{Information Environment}
\label{app:models_environment}

\textbf{News corpus and effective dates.} We download WARC archives from CC-News and apply a seven-stage quality filter based on FineWeb \citep{penedo2024fineweb}. The pipeline combines text cleaning, FastText language identification, Gopher repetition and quality heuristics, C4 rules, and a FineWeb quality pass. We then deduplicate the aggregate corpus by URL, title, and MinHash, rather than deduplicating each file separately. We embed the retained articles and store them in daily FAISS shards. The index used in the reported experiments contains 516 daily shards spanning February 2025 to July 2026. We embed with Qwen3-Embedding-8B and search an inner-product index; a query returns ten candidate articles by default, although the agent may request more.

We assign each article an effective date by taking the latest date from its structured markup, body, and URL, substituting the crawl date when the body date is missing. This conservative rule can move an article to a later date, reducing the risk that information becomes available to an instance prematurely. Nevertheless, historical availability cannot be reconstructed perfectly from web data: publication timestamps are not always complete or consistently represented across sources, and some pages may be updated after their recorded dates. We therefore treat the effective date as a conservative approximation of when an article could have been available.

\textbf{Documents offered.} For target quarter $q$, we direct the agent to the $q{-}1$ earnings-call transcript and the $q{-}1$ and $q{-}4$ filings. We also suggest the $q{-}2$ and $q{-}3$ filings as optional reading on trends and seasonality. The agent can request other filings or transcripts published on or before $\tau$, but the server blocks the target-quarter filing and call and any document dated after $\tau$. We expose filings through a table of contents so agents can read selected sections of a long filing.

\subsection{Tools and Run Protocol}
\label{app:tools_protocol}

\textbf{Tool suite.} The MCP server of Section~\ref{sec:data} exposes six tools:

\begin{itemize}
    \item \texttt{list\_financial\_report\_sections} / \texttt{read\_financial\_report\_section}: enumerate and read individual sections of historical filings (balance sheet, MD\&A, segment tables), so that long documents can be navigated selectively instead of loaded whole.
    \item \texttt{read\_earnings\_call\_transcript}: retrieve the transcript of a past earnings call, where management discusses the outlook for coming quarters.
    \item \texttt{financial\_search}: semantic retrieval over the date-sharded news index, subject to the temporal filtering described in Appendix~\ref{app:models_environment}.
    \item \texttt{read\_news\_article}: fetch the full text of a retrieved article.
    \item \texttt{financial\_calculate}: compute the income statement and EPS from the agent's driver values. The calculator applies these inputs to the schema and executes the accounting DAG in topological order.
\end{itemize}

\textbf{Run protocol.} Each run is a single attempt that ends when the agent sends its final message after calling \texttt{financial\_calculate}. We impose no turn limit beyond the model's context window and do not retry runs. When the agent requests a missing document or one dated after $\tau$, the server returns an error and the run continues. We log and count these refusals. If we cannot parse the final message into the required JSON, we record a format failure and no forecast. These rules apply to all models, harnesses, and information conditions, and failures are reported alongside point forecast error, as Section~\ref{sec:eval} describes.

\subsection{Evaluation and Aggregation}
\label{app:evaluation_details}

\textbf{Interval score.} For a reported p10 $l$, p90 $u$, and reported EPS $y$, the interval score of \citet{gneiting_strictly_2007} for a central $(1-\alpha)$ interval with $\alpha=0.2$ is
$$\mathrm{IS} = \frac{1}{P}\Big[(u - l) + \tfrac{2}{\alpha}\,(l - y)^{+} + \tfrac{2}{\alpha}\,(y - u)^{+}\Big],$$
where $(x)^{+}=\max(x,0)$ and $P$ is the share price used for PDE, so IS is in the same price-deflated units as PDE, with lower scores indicating better performance. The score penalizes each miss by ten times its distance from the nearest bound and charges for interval width. It is proper for the elicited p10 and p90 quantiles, balancing the penalties for missed outcomes against the cost of wider intervals. We report the mean over runs that return an interval because propriety holds for the expected score. A median would fall among covered runs, where the score equals the width, and would therefore reward narrow intervals.

\textbf{Efficiency measures.} We record model requests, tool calls, output and reasoning tokens, and total input tokens for each run. Total input sums the tokens carried by every request, including cached tokens and conversation history sent again in later requests. We report medians over runs and use token counts because providers price cached input differently. Format compliance is the share of runs whose final message parses into the required forecast.

\textbf{Company-level resampling.} Two quarters from the same company may not be independent. Resampling companies instead of instances, and keeping all of a company's instances together, moves the endpoints of the intervals against consensus in Table~\ref{tab:main} by at most 0.002 pp and changes no conclusion.

\section{Companies in the Benchmark}
\label{app:companies}

Table~\ref{tab:companies} lists the 86 companies by sector. To describe the sample, we use sector labels and market capitalizations from Yahoo Finance as of September 2026. The median market capitalization is \$212B, and 85 of the 86 companies exceed \$10B.

\begin{table}[!htb]
\caption{\textbf{Benchmark companies by sector.} The 86 companies across eight sectors, with company counts ($n$) and ticker symbols. Sector classifications follow Yahoo Finance as of September 2026.}
\label{tab:companies}
\begin{center}
\small
\setlength{\tabcolsep}{4pt}
\begin{tabular}{@{}p{3.2cm} r p{9.3cm}@{}}
\toprule
Sector & $n$ & Tickers \\
\midrule
    Technology & 20 & ADBE, AMAT, AMD, AVGO, CRM, CRWD, CSCO, DELL, IBM, INTC, INTU, MSFT, MU, NVDA, ORCL, PANW, PLTR, QCOM, SMCI, UBER \\
    Consumer Cyclical & 15 & AMZN, BBY, CCL, CMG, CVNA, GM, HD, LOW, LULU, M, MCD, MELI, NKE, SBUX, TSLA \\
    Financial Services & 14 & AXP, BAC, BLK, C, COIN, GS, HOOD, JPM, MA, MS, PYPL, SPGI, V, WFC \\
    Industrials & 11 & BA, CAT, DAL, GE, HON, LMT, PLUG, RTX, UAL, UNP, UPS \\
    Healthcare & 9 & ABBV, AMGN, ISRG, JNJ, LLY, MRK, PFE, TMO, UNH \\
    Consumer Defensive & 7 & COST, KHC, KO, PEP, PG, TGT, WMT \\
    Communication Services & 6 & CMCSA, DIS, GOOGL, META, NFLX, WBD \\
    Energy & 4 & COP, CVX, SLB, XOM \\
\bottomrule
\end{tabular}
\end{center}
\end{table}

\section{Model Details}
\label{app:models}
Table~\ref{tab:models} summarizes the models evaluated in our experiments.
We report release dates and knowledge cutoffs from vendor disclosures where
available; ``undisclosed'' indicates that the vendor has not disclosed a value
for that field. Four models have published knowledge cutoffs preceding
all benchmark announcements. Qwen 3.5 and GLM 4.5 Air do not disclose knowledge
cutoffs.
\begin{table}[!htb]
\caption{\textbf{Evaluated model specifications.} Model identifiers, total and active parameter counts, release dates, and knowledge cutoffs from vendor disclosures. Entries marked ``undisclosed'' were not provided by the vendor.}
\label{tab:models}
\begin{center}
\small
\setlength{\tabcolsep}{4pt}
\begin{tabular}{@{}l l l l@{}}
\toprule
Model identifier & \shortstack{Parameters\\total / active} & Released & \shortstack{Knowledge\\cutoff} \\
\midrule
claude-opus-4.6 & undisclosed & 2026-02-05 & 2025-08 \\
gpt-5.5 & undisclosed & 2026-04-23 & 2025-12-01 \\
gpt-5.4 & undisclosed & 2026-03-05 & 2025-08-31 \\
gpt-5-mini & undisclosed & 2025-08-07 & 2024-05-31 \\
qwen3.5-122b-a10b & 122B / 10B & 2026-02-24 & undisclosed \\
glm-4.5-air & 106B / 12B & 2025-07-28 & undisclosed \\
\bottomrule
\end{tabular}
\end{center}
\end{table}

\section{Harness Settings}
\label{app:harness}

An agent harness is the framework between the model and the tools: it assembles the prompt sent at each turn, routes the model's tool calls to the MCP server, and decides what to keep in the context as the conversation grows. The harness experiments in Section~\ref{sec:harness} hold everything else fixed, namely the 149 instances, the schemas, the MCP server and its six tools described in Appendix~\ref{app:tools_protocol}, and the system and user prompts given in Appendix~\ref{app:prompts}, and vary only the harness and the reasoning effort.

Table~\ref{tab:harness_settings} lists the three harnesses. OpenCode is the primary harness for all six models. We drive it through its SDK and disable its built-in shell, file, and web tools, so the six MCP tools are the agent's only interface. Claude Code, Anthropic's agent harness, runs Opus 4.6 under the same restriction. The vanilla harness is a plain tool-calling loop over the OpenAI SDK that we wrote for this comparison. It resends the full conversation each turn and adds no system prompt beyond the benchmark's, no context management, and no tool orchestration, so it shows how GPT-5.5 behaves without the scaffolding the other harnesses provide.

\begin{table}[!htb]
\caption{\textbf{Harness configurations.} Every harness runs at the provider's default reasoning effort; the maximum-effort configurations for Opus 4.6 and GPT-5.5 are listed in Table~\ref{tab:harness}. Temperature is left unset throughout.}
\label{tab:harness_settings}
\begin{center}
\small
\setlength{\tabcolsep}{4pt}
\begin{tabular}{@{}p{2cm} p{2cm} p{1.8cm} p{7.3cm}@{}}
\toprule
Harness & Version & Models & Configuration \\
\midrule
OpenCode & SDK 1.17.9 & all six & Driven through the SDK; built-in shell, file, and web tools disabled by the agent definition. Primary harness. \\
Claude Code & 2.1.269 & Opus 4.6 & Driven through the SDK; built-in shell, file, and web tools disabled. \\
Vanilla harness & ours & GPT-5.5 & Plain tool-calling loop over the OpenAI SDK; resends the full conversation every turn; no system prompt beyond Listing~\ref{lst:system_prompt}, no context management, no tool orchestration. \\
\bottomrule
\end{tabular}
\end{center}
\end{table}

All harnesses send the same benchmark prompts and call the same MCP server. Every model runs at the provider's default reasoning effort. To separate the effect of effort from that of the harness, we also run Opus 4.6 at maximum effort under both OpenCode and Claude Code, and GPT-5.5 at maximum effort under OpenCode. No configuration sets a temperature.

\section{Memorization Test for Qwen 3.5}
\label{app:probe}

The vendor released Qwen 3.5 on 24 February 2026, during the benchmark window, without publishing a knowledge cutoff. Of the 149 instances, 79 have announcements after that date and 70 on or before it. We therefore probe whether the model can recall reported EPS across successive announcement periods, including the benchmark window.

\textbf{Probe.} We ask the model for a named company's GAAP diluted EPS in a specified fiscal quarter, without tools or documents, and instruct it to answer UNKNOWN if it does not know. We use an unguided prompt that asks directly and a guided prompt that encourages recall from training data. We query each firm-quarter three times with each prompt and count an answer as a \textit{hit} when its PDE against reported EPS, defined in Section~\ref{sec:eval}, is at most 0.02\%.

To assess incidental matches, we compare four strategies that use no memory of the target quarter at three candidate thresholds, which Table~\ref{tab:probe_null} reports. Copying analyst consensus, the strongest tested strategy at 0.02\%, achieves a 9\% hit rate; its rate rises to 22\% at 0.05\%. We use the tighter threshold to reduce such matches, while recognizing that a hit need not imply memorization. Each reference strategy receives one attempt per firm-quarter, compared with six for the model.

\begin{table}[!htb]
\caption{\textbf{Reference hit rates for the memorization test.} Hit rates of non-memorizing strategies on 149 firm-quarters at three PDE thresholds. A hit occurs when the forecast's PDE does not exceed the threshold.}
\label{tab:probe_null}
\begin{center}
\small
\setlength{\tabcolsep}{4pt}
\begin{tabular}{l r r r}
\toprule
Strategy & 0.02\% & 0.05\% & 0.10\% \\
\midrule
Copy analyst consensus & 9\% & 22\% & 37\% \\
Copy previous quarter ($q{-}1$) & 4\% & 14\% & 27\% \\
Copy same quarter last year ($q{-}4$) & 4\% & 8\% & 22\% \\
Guess zero & 1\% & 2\% & 5\% \\
\bottomrule
\end{tabular}
\end{center}
\end{table}

\textbf{Firm-quarters probed.} We construct a historical sample from the same 86 companies, using announcements from June 2024 to June 2026. We group these into nine calendar-quarter buckets, from 2024Q2 to 2026Q2, and sample 30 firm-quarters per bucket, using announcement dates estimated at sampling time for some older quarters. The first bucket contains only seven available cases, giving 247 in total. After we replaced the estimated dates with actual announcement dates from IBES, one firm-quarter moved from the third bucket to the fourth, so Table~\ref{tab:probe} lists 29 and 31 for these two periods. This sample lets us compare elicited recall across announcement periods. We also probe all 149 benchmark instances and report the results in Table~\ref{tab:probe}.

\textbf{Results.} We count a firm-quarter as a hit if at least one of its six responses meets the threshold. Table~\ref{tab:probe} shows hits for announcements in 2024 and none from January 2025 onward, including every benchmark instance. Across six responses per instance, none of the 149 benchmark instances yields a hit at the selected threshold, indicating no detectable recall under the tested protocol. When asked about its own cutoff, the model answers ``December 2024''. This response is consistent with the observed pattern but provides only weak additional evidence.

\begin{table}[!htb]
\caption{\textbf{Qwen 3.5 memorization test.} Recall results by announcement period and on the benchmark instances. A firm-quarter counts as a hit if at least one of six responses achieves PDE $\leq 0.02\%$.}
\label{tab:probe}
\begin{center}
\small
\setlength{\tabcolsep}{4pt}
\begin{tabular}{l r r r}
\toprule
Announcement dates & Firm-quarters & Hits & Hit rate \\
\midrule
June 2024 & 7 & 1 & 14\% \\
11 July to 9 September 2024 & 29 & 10 & 34\% \\
1 October to 19 December 2024 & 31 & 6 & 19\% \\
10 January 2025 to June 2026 (six buckets) & 180 & 0 & 0\% \\
\midrule
Benchmark instances (December 2025 to June 2026) & 149 & 0 & 0\% \\
\bottomrule
\end{tabular}
\end{center}
\end{table}

\section{Additional Results}
\label{app:results}

\textbf{Detailed evaluation results.}
Tables~\ref{tab:calib}--\ref{tab:pairwise} supplement the evaluation in Section~\ref{sec:results}. Table~\ref{tab:calib} reports interval coverage, tail rates, width, and score. These statistics include only runs that return an interval, with the denominator reported for each model. Table~\ref{tab:eff} reports resource use, format compliance, and the number of instances with large errors. Peak context is the largest input-token count in any single request within a run. Table~\ref{tab:pairwise} reports all pairwise model comparisons underlying the point-forecast analysis in Section~\ref{sec:point}.

\textbf{Resource use and compliance.}
The resource measurements show that processing more input does not necessarily accompany lower error. The three leading models make 35--49 tool calls per run, compared with 20--27 for the others, but GPT-5-mini processes the most input at 1.35M tokens per run. Its 28 model requests repeatedly carry the conversation history, whereas Table~\ref{tab:eff} shows 12--21 requests for the other models. Opus 4.6 and GPT-5.5 therefore achieve lower error while processing fewer input tokens than GPT-5-mini. All models return parseable forecasts in at least 98.2\% of runs. Eight of the nine unparseable runs come from GPT-5-mini, while GLM 4.5 Air, the model with the highest error, parses every run, so compliance does not track point forecast error.

\begin{table}[!htb]
\caption{\textbf{Prediction interval calibration and forecast bias.} Interval metrics pool runs returning a p10--p90 interval. Below, Inside, and Above report outcome percentages relative to the interval; Inside is empirical coverage. Width is the median interval width, and Score is the mean interval score. Signed error is the median of $(y-\hat y)/P$ over scoreable runs, with a 90\% bootstrap confidence interval that resamples instances; positive values indicate underprediction.}
\label{tab:calib}
\begin{center}
\small
\setlength{\tabcolsep}{5pt}
\begin{tabular}{@{}l r rrr rr l@{}}
\toprule
 & & \multicolumn{3}{c}{Outcome location (\%)} & & & \\
\cmidrule(lr){3-5}
Model & \shortstack{Intervals\\returned} & Below & Inside & Above & \shortstack{Width\\(pp)} & \shortstack{Score\\(pp)} & \shortstack{Signed error, pp\\(90\% CI)} \\
\midrule
Opus 4.6 & 447 / 447 & 10 & 70 & 20 & 0.41 & 2.87 & $+0.021$ $[+0.007,+0.044]$ \\
GPT-5.5 & 447 / 447 & 10 & 61 & 29 & 0.33 & 3.08 & $+0.038$ $[+0.016,+0.068]$ \\
GPT-5.4 & 447 / 447 & 16 & 52 & 32 & 0.25 & 4.29 & $+0.037$ $[+0.015,+0.061]$ \\
GPT-5-mini & 439 / 447 & 18 & 50 & 32 & 0.39 & 19.61 & $+0.047$ $[+0.010,+0.086]$ \\
Qwen 3.5 & 446 / 447 & 25 & 37 & 38 & 0.26 & 4.87 & $+0.039$ $[+0.006,+0.072]$ \\
GLM 4.5 Air & 409 / 447\footnotemark[1] & 22 & 44 & 34 & 0.33 & 6.48 & $+0.048$ $[+0.021,+0.080]$ \\
\midrule
Analyst consensus & -- & -- & -- & -- & -- & -- & $+0.028$ $[+0.008,+0.051]$ \\
Expected split & -- & 10 & 80 & 10 & -- & -- & -- \\
\bottomrule
\end{tabular}
\end{center}
\end{table}
\footnotetext[1]{Context compaction inside the harness was triggered on 49 of GLM 4.5 Air's 447 runs (its 128K window); these runs are scored as delivered and account for all of its missing intervals.}

\begin{table}[!htb]
\caption{\textbf{Resource use, format compliance, and large errors.} Panel (a) reports medians over runs; input includes cached tokens. The APIs for Opus 4.6 and GLM 4.5 Air do not report reasoning tokens separately, so their reasoning is counted in Output and the Reasoning column is n/a. Panel (b) reports format compliance, the share of runs whose final message parses into the required forecast, and the number of the 149 instances whose median PDE over runs exceeds 1\%.}
\label{tab:eff}
\begin{center}
\small
\setlength{\tabcolsep}{5pt}
\begin{tabular}{@{}l rr rrrr@{}}
\toprule
\multicolumn{7}{@{}l}{\textbf{(a) Resource use per run}} \\
Model & Requests & Tools & \shortstack{Input\\(K)} & \shortstack{Output\\(K)} & \shortstack{Reasoning\\(K)} & \shortstack{Peak context\\(K)} \\
\midrule
Opus 4.6 & 15 & 35 & 1056 & 9.7 & n/a & 112 \\
GPT-5.5 & 12 & 47 & 837 & 3.6 & 2.6 & 119 \\
GPT-5.4 & 13 & 49 & 949 & 3.7 & 7.2 & 125 \\
GPT-5-mini & 28 & 27 & 1354 & 2.7 & 3.6 & 90 \\
Qwen 3.5 & 14 & 23 & 542 & 3.0 & 5.1 & 61 \\
GLM 4.5 Air & 21 & 20 & 738 & 3.4 & n/a & 59 \\
\bottomrule
\end{tabular}
\par\vspace{7pt}
\begin{tabular}{@{}l rr@{}}
\toprule
\multicolumn{3}{@{}l}{\textbf{(b) Format compliance and error frequency}} \\
Model & Compliance (\%) & \shortstack{Instances with PDE $>1\%$} \\
\midrule
Opus 4.6 & 100 & 8 \\
GPT-5.5 & 100 & 11 \\
GPT-5.4 & 100 & 11 \\
GPT-5-mini & 98.2 & 13 \\
Qwen 3.5 & 99.8 & 14 \\
GLM 4.5 Air & 100 & 23 \\
\bottomrule
\end{tabular}
\end{center}
\end{table}

\textbf{Source-condition interaction.}
To complement the information-source comparisons in Section~\ref{sec:channels} and Table~\ref{tab:info_ablation}, we compute an interaction estimate for each instance. Let $d_{\mathrm{N}}$, $d_{\mathrm{F}}$, and $d_{\mathrm{B}}$ be the changes in the instance's PDE, the median over its three runs, when news, filings and transcripts, or both are removed, relative to the full environment. The interaction estimate is $d_{\mathrm{B}}-(d_{\mathrm{N}}+d_{\mathrm{F}})$, which is positive when removing both sources hurts more than the two single removals combined. For GPT-5.5, its median over the 149 instances is $+0.037$ pp, with a bootstrap 90\% interval of $[+0.020,+0.066]$ computed as in Section~\ref{sec:eval}, and it is positive on 100 instances. A median of instance-level differences need not equal the difference between the medians in Table~\ref{tab:info_ablation}, so this estimate is not the gap between the displayed values.

\textbf{Source-removal groups.} For each instance and condition, we take the median PDE over the three runs; a removal effect is the difference between this value in the reduced condition and in the full environment. An effect is harmful when it exceeds 0.057 pp, the median over the 149 instances and four conditions of the range between the highest and lowest PDE among an instance's three runs. Table~\ref{tab:source_groups} places an instance in the first group when no removal is harmful, in the second when only the joint removal is harmful, and in the third when at least one single removal is harmful. The rank correlation of $-0.01$ is computed over all 149 instances between the two single-removal effects.

\begin{table}[!htb]
\caption{\textbf{Instance-level effects of information-source removal.} GPT-5.5 results on 149 firm-quarters. A removal is harmful when its error increase exceeds 0.057 pp, the median spread of PDE across an instance's three runs. Error increases from removing filings and transcripts versus removing news have a Spearman correlation of $-0.01$.}
\label{tab:source_groups}
\begin{center}
\small
\setlength{\tabcolsep}{5pt}
\begin{tabular}{@{}l r r@{}}
\toprule
Group & Instances & Share (\%) \\
\midrule
No removal is harmful & 67 & 45 \\
Only removing both is harmful & 32 & 21 \\
A single removal is harmful & 50 & 34 \\
\quad only removing filings and transcripts & 17 & 11 \\
\quad only removing news & 24 & 16 \\
\quad either removal & 9 & 6 \\
\bottomrule
\end{tabular}
\end{center}
\end{table}

\begin{table}[!htb]
\caption{\textbf{Pairwise model comparisons of point forecast error.} Median paired $\Delta$PDE (A $-$ B) on 149 firm-quarters, with 90\% bootstrap confidence intervals and counts of instances where each model has lower PDE. Negative differences indicate lower error for model A.}
\label{tab:pairwise}
\begin{center}
\small
\setlength{\tabcolsep}{4pt}
\begin{tabular}{l l r l r}
\toprule
Model A & Model B & \shortstack{$\Delta$PDE\\(A $-$ B), pp} & 90\% CI & \shortstack{A wins /\\B wins} \\
\midrule
Opus 4.6 & GPT-5.5 & $-0.004$ & $[-0.013, +0.002]$ & 81 / 68 \\
Opus 4.6 & GPT-5.4 & $-0.001$ & $[-0.009, +0.005]$ & 79 / 70 \\
Opus 4.6 & GPT-5-mini & $-0.032$ & $[-0.042, -0.026]$ & 105 / 44 \\
Opus 4.6 & Qwen 3.5 & $-0.045$ & $[-0.054, -0.031]$ & 110 / 39 \\
Opus 4.6 & GLM 4.5 Air & $-0.048$ & $[-0.063, -0.031]$ & 107 / 42 \\
GPT-5.5 & GPT-5.4 & $+0.002$ & $[-0.000, +0.006]$ & 67 / 82 \\
GPT-5.5 & GPT-5-mini & $-0.029$ & $[-0.038, -0.020]$ & 107 / 42 \\
GPT-5.5 & Qwen 3.5 & $-0.035$ & $[-0.046, -0.019]$ & 105 / 44 \\
GPT-5.5 & GLM 4.5 Air & $-0.042$ & $[-0.063, -0.032]$ & 105 / 44 \\
GPT-5.4 & GPT-5-mini & $-0.028$ & $[-0.042, -0.019]$ & 100 / 49 \\
GPT-5.4 & Qwen 3.5 & $-0.038$ & $[-0.057, -0.021]$ & 108 / 41 \\
GPT-5.4 & GLM 4.5 Air & $-0.038$ & $[-0.063, -0.026]$ & 106 / 43 \\
GPT-5-mini & Qwen 3.5 & $-0.007$ & $[-0.017, +0.003]$ & 83 / 66 \\
GPT-5-mini & GLM 4.5 Air & $-0.009$ & $[-0.016, +0.002]$ & 83 / 66 \\
Qwen 3.5 & GLM 4.5 Air & $-0.003$ & $[-0.015, +0.008]$ & 77 / 72 \\
\bottomrule
\end{tabular}
\end{center}
\end{table}

\textbf{Accounting-stage decomposition.}
To complement the analysis in Section~\ref{sec:departure}, we decompose
forecast error on the 125 instances with an operating-income subtotal.
For each run, we measure absolute forecast error at operating income,
pretax income, and EPS, placing all three on the PDE scale by dividing the
two income errors by the target quarter's reported diluted share count and
then by share price $P$. Writing $e_{\mathrm{op}}$, $e_{\mathrm{pt}}$, and
$e_{\mathrm{EPS}}$ for these errors, the three components are
$e_{\mathrm{op}}$, $e_{\mathrm{pt}}-e_{\mathrm{op}}$, and
$e_{\mathrm{EPS}}-e_{\mathrm{pt}}$. Thus they sum to PDE, and a negative
increment means that error shrinks at that stage. We take the median of each
component over an instance's three runs and report medians across instances.
The last column of Table~\ref{tab:layers} counts the instances on which each
component has the largest absolute magnitude.

\textbf{Evidence coding for Table~\ref{tab:departure}.} For each run on a far instance, we reset each driver in turn to its year-ago value, with zero growth or margin change for growth and margin drivers, recompute EPS with the calculator, and take the driver whose reset changes EPS the most as the run's most-adjusted driver. We then read that driver's evidence line in the agent's final message and count the run as citing a specific number if the part of the line before any arrow to the computed value contains a dollar amount, a percentage, a figure with thousands separators, or a number with a unit such as basis points, million, or billion. A run whose most-adjusted driver has no evidence line counts as not citing a number, so the denominator is all runs on the 50 far instances.

\section{Case Study: Boeing's Divestiture}
\label{app:boeing_case}

Boeing's fourth quarter of 2025 shows how the driver records locate differences between forecasts. Boeing completed the \$10.55B sale of Jeppesen and ForeFlight on 3 November 2025, about twelve weeks before its earnings announcement. The sale price is not earnings; the forecasting question is how large a gain Boeing would recognize in the quarter. Boeing's schema has a driver for this gain, which enters operating income. In every run, Opus 4.6, GPT-5.5, and GPT-5.4 entered \$8.0--9.6B, working from the sale price and the net book value of the assets held for sale disclosed in the third-quarter filing; with their other drivers unchanged, this entry adds \$9.0--12.4 to their EPS forecasts. Qwen 3.5 found the transaction but concluded that no gain would be recognized in the quarter, and it and GLM 4.5 Air entered a gain close to zero in every run. GPT-5-mini entered about \$9.6B in two of its three runs, but its other drivers put operating income \$4--5B below the leading models' forecasts. Boeing reported GAAP diluted EPS of \$10.23, against a consensus of $-\$0.31$. The median forecasts over three runs were \$8.78 for Opus 4.6, \$9.16 for GPT-5.5, \$10.48 for GPT-5.4, \$5.06 for GPT-5-mini, $-\$7.44$ for Qwen 3.5, and $-\$7.57$ for GLM 4.5 Air. The gain driver accounts for \$10--12 of the roughly \$17 gap between the leading models and Qwen 3.5 and GLM 4.5 Air, and the records attribute the rest to revenue and cost drivers.

\begin{table}[!htb]
\caption{\textbf{Forecast error decomposition by accounting stage.} Results on 125 instances with an operating-income subtotal. Operating reports the median error at operating income; Non-operating and Tax and shares report the median error increments from operating to pretax income and from pretax income to EPS, respectively, all on the PDE scale. Negative increments indicate a reduction in error. The last column counts instances where each component is largest in absolute magnitude (O: operating; N: non-operating; T: tax and shares).}
\label{tab:layers}
\begin{center}
\small
\setlength{\tabcolsep}{4pt}
\begin{tabular}{l r r r r}
\toprule
Model & \shortstack{Operating \\(pp)} & \shortstack{Non-operating \\(pp)} & \shortstack{Tax and shares \\(pp)} & \shortstack{Largest component \\ O / N / T} \\
\midrule
Opus 4.6 & 0.092 & 0.007 & $-0.017$ & 80 / 26 / 19 \\
GPT-5.5 & 0.089 & 0.013 & $-0.016$ & 83 / 20 / 22 \\
GPT-5.4 & 0.098 & 0.013 & $-0.018$ & 88 / 22 / 15 \\
GPT-5-mini & 0.149 & 0.020 & $-0.031$ & 84 / 27 / 14 \\
Qwen 3.5 & 0.166 & 0.012 & $-0.025$ & 91 / 23 / 11 \\
GLM 4.5 Air & 0.166 & 0.012 & $-0.032$ & 86 / 26 / 13 \\
\bottomrule
\end{tabular}
\end{center}
\end{table}

\section{Complete Example of a Generated Schema}
\label{appendix:schema}

Listing~\ref{lst:full_schema} reproduces the complete, unedited schema that our extraction pipeline generated for Microsoft (\texttt{MSFT}), fiscal 2026Q3, following Section~\ref{sec:offline_extraction}. The \texttt{variables\_baseline} block stores leaf values for 2025Q3, the year-ago quarter, while \texttt{formulas} defines the DAG and \texttt{prediction\_schema} lists the agent's input fields. A field can set several leaves: \texttt{cogs\_margin\_delta\_bps}, for example, controls both cost-of-revenue lines.

This schema represents income tax as an absolute value, \texttt{income\_taxes\_abs}, with the year-ago expense as its default. The share-count field carries the three references described in Appendix~\ref{app:schema_construction}: \texttt{last\_q}, \texttt{ttm}, and \texttt{q4}. We use the reported EPS values in \texttt{\_validation} to verify the schema and score forecasts, and, as Section~\ref{sec:data} describes, keep this block inaccessible to the agent.

\begin{lstlisting}[language=json, caption={The complete schema \texttt{MSFT\_2026Q3.json}.}, label={lst:full_schema}]
{
  "ticker": "MSFT",
  "target_fq": "2026Q3",
  "baseline_fq": "2025Q3",
  "variables_baseline": {
    "product_revenue": 15319.0,
    "service_and_other_revenue": 54747.0,
    "product_cogs": 3037.0,
    "service_and_other_cogs": 18882.0,
    "rd": 8198.0,
    "sales_and_marketing": 6212.0,
    "general_and_administrative": 1737.0,
    "other_expense_net": 623.0,
    "income_taxes": 5553.0,
    "shares": 7461.0
  },
  "formulas": {
    "total_revenue": "product_revenue + service_and_other_revenue",
    "total_cogs": "product_cogs + service_and_other_cogs",
    "gross_profit": "total_revenue - total_cogs",
    "total_opex": "rd + sales_and_marketing + general_and_administrative",
    "operating_income": "gross_profit - total_opex",
    "pretax_income": "operating_income - other_expense_net",
    "net_income": "pretax_income - income_taxes",
    "earnings_to_common": "net_income",
    "eps": "earnings_to_common / shares"
  },
  "prediction_schema": {
    "product_revenue_yoy": {
      "variable": "product_revenue",
      "type": "yoy",
      "constituents": [
        "product_revenue"
      ],
      "description": "Product revenue YoY growth rate vs Q-4"
    },
    "service_and_other_revenue_yoy": {
      "variable": "service_and_other_revenue",
      "type": "yoy",
      "constituents": [
        "service_and_other_revenue"
      ],
      "description": "Service and other revenue YoY growth rate vs Q-4"
    },
    "cogs_margin_delta_bps": {
      "variable": "total_cogs",
      "type": "bps",
      "anchor": "total_revenue",
      "constituents": [
        "product_cogs",
        "service_and_other_cogs"
      ],
      "description": "Combined cost of revenue as % of total revenue, delta vs Q-4"
    },
    "rd_yoy": {
      "variable": "rd",
      "type": "yoy",
      "constituents": [
        "rd"
      ],
      "description": "R&D expense YoY growth rate vs Q-4"
    },
    "sales_and_marketing_yoy": {
      "variable": "sales_and_marketing",
      "type": "yoy",
      "constituents": [
        "sales_and_marketing"
      ],
      "description": "Sales and marketing expense YoY growth rate vs Q-4"
    },
    "general_and_administrative_yoy": {
      "variable": "general_and_administrative",
      "type": "yoy",
      "constituents": [
        "general_and_administrative"
      ],
      "description": "General and administrative expense YoY growth rate vs Q-4"
    },
    "other_expense_net_abs": {
      "variable": "other_expense_net",
      "type": "absolute",
      "description": "other_expense_net absolute dollar value for the target quarter (predict the amount directly; 0 if none expected)"
    },
    "income_taxes_abs": {
      "variable": "income_taxes",
      "type": "absolute",
      "default": 5553.0,
      "description": "income_taxes absolute dollar value for the target quarter (below-pretax line; predict directly, 0 if none expected)."
    },
    "shares_abs": {
      "variable": "shares",
      "type": "absolute",
      "default": 7461.0,
      "references": {
        "last_q": 7460.0,
        "ttm": 7462.0,
        "q4": 7461.0
      },
      "description": "shares: diluted weighted-average share count for the target quarter (predict the count directly; drifts slowly with buybacks/issuance)."
    }
  },
  "_validation": {
    "baseline_fq_eps": 3.46,
    "target_fq_eps": 4.27,
    "target_fq_pretax": 39340.0
  }
}
\end{lstlisting}

\section{Prompts}
\label{app:prompts}

For a given instance and information condition, all models and harnesses receive the same system and user prompts. The system prompt in Listing~\ref{lst:system_prompt} is static. The user prompt is generated per instance from the schema and the cutoff; Listing~\ref{lst:user_prompt} is the one issued for Microsoft, fiscal 2026Q3, the instance whose schema Appendix~\ref{appendix:schema} shows. The reduced information conditions of Section~\ref{sec:channels} change only the text that refers to removed tools. When news is removed, the news tools are not registered, and the user prompt drops the news window and the instruction to search news; the system prompt is unchanged, and no run attempts a news call. When filings and transcripts are removed, the three document tools are not registered, the user prompt replaces the required reading with a news search and names the year-ago baseline and the tax and share references as the only historical figures, and the system prompt states that the document tools are deliberately unavailable. When both are removed, no research tool is registered, and both prompts tell the agent to work from the year-ago baseline and the references alone. The release includes the prompts for every condition.

\begin{lstlisting}[style=prompt, caption={The system prompt.}, label={lst:system_prompt}]
# Financial Forecaster (Fundamental Update)

Run ONE complete forecasting iteration for a specified target ticker and fiscal quarter. Your task is to act as a quantitative analyst updating a financial model. Start by carefully reading the user message to identify the exact variables and units you must (and can only) predict, and to obtain the target ticker, fiscal quarter, and the exact Q-4 Baseline Data. From there, use the available tools at your own discretion — in whatever order and combination makes sense — to read the baseline filings and recent qualitative text (Reports, ECT, News), predict the relative changes for specific core drivers, and offload the actual computation entirely to the calculator tool.

## CRITICAL CONSTRAINTS

- **No Numerical Hallucination:** You MUST NOT predict absolute values for `Total Revenue` or `EPS` directly. You ONLY predict the relative changes (YoY growth rates, margin bps deltas) for the exact fields requested in the prompt.
- **No Manual Calculation:** Do not attempt to compute the final absolute values yourself. You MUST use the `financial_calculate` tool.
- **Strict Input Fields:** When calling `financial_calculate`, the arguments must strictly match the required prediction fields (e.g., `products_revenue_yoy`). Never invent or hallucinate new parameters.
- **Maximize Information Retrieval:** Before predicting, you MUST maximize your use of the search tool and read multiple relevant news articles. Do not jump to conclusions after reading only one or two articles.
- **Comprehensive News Coverage Before Predicting:** Do not call `financial_calculate` until you have built a comprehensive, multi-dimensional picture of the company's recent situation. This means actively searching across multiple angles — revenue trends by segment, margin pressures, macro and regulatory factors, competitive dynamics, and management signals. Only when further reading would no longer materially change your estimates should you stop and move to prediction.
- **Recency Bias:** When reading news, assign higher analytical weight to articles published closer to the cutoff date.
- **No Control Over the Information Cutoff:** The news search tool is restricted server-side to a fixed information window for this run. There is no parameter to widen it, query past it, or ask for a different date. Do not assume a "current date" — rely only on what the tools return.
- **Never Fabricate on Tool Failure:** If any tool errors, times out, or becomes unavailable — including `financial_calculate` — do not retry indefinitely and do not invent a final JSON from memory. After a few failed attempts, output a single line starting with `ERROR:` describing what failed, and stop.
- **No Reverse-Engineering From Consensus:** If you encounter analyst consensus estimates for total EPS or total revenue (e.g. in an earnings-preview article), do NOT treat that aggregate figure as a target and back-solve your driver values to match it. Build each driver estimate independently and bottom-up from segment-specific evidence (unit sales, pricing, regional trends, cost data) gathered during your research. You may use a consensus figure only as a sanity check AFTER you have already derived your estimates — a large gap from consensus is not, by itself, evidence that your estimate is wrong, as long as it is grounded in specific evidence you found. This is enforced procedurally: see the mandatory per-field evidence block in OUTPUT FORMAT below — every field's value must trace to its own cited evidence, not to a total you picked first.

### Anti-Hallucination Rules

The most common failure mode is attempting to act as a calculator or hallucinating variables that do not exist in the schema.
- If the target requires `products_revenue_yoy`, NEVER attempt to provide `total_revenue_yoy`.
- If a qualitative news article mentions a $1 billion fine or a one-time gain, but there is no corresponding node in the prediction list, NEVER attempt to manually adjust any numbers to deduct or add it.
- Trust that the Python backend will use Kahn's Algorithm for correct topological sorting. Fully trust the calculated results summary returned by the tool.

## RESEARCH TOOLS

You have two categories of information sources, available in any order and combination you see fit:

- **Filings:** `read_earnings_call_transcript` and `list_financial_report_sections` -> `read_financial_report_section`, covering the Q-1 and Q-4 quarters, for historical performance, segment breakdown, and management guidance. Large sections are returned in chunks — if a response has `"has_more": true`, call `read_financial_report_section` again with `offset` set to the returned `next_offset` to keep reading the same section; do not treat a single chunked call as the complete section.
- **News:** `financial_search` with distinct queries covering different angles — revenue outlook by segment, cost pressures/tariffs/supply chain, macro and regulatory developments, competitive dynamics — followed by `read_news_article` on the most relevant results.

There is no required reading order or fixed number of calls. Use your judgment about what to read and when, subject only to the constraints above (comprehensive coverage before predicting, recency weighting, and never fabricating on tool failure).

## INTERFACE DEFINITIONS

**Available Tools** (ticker and fiscal quarter for this run are fixed by the environment — do not pass them):
- `read_earnings_call_transcript(quarter)`
- `list_financial_report_sections(quarter)`
- `read_financial_report_section(quarter, section_name)`
- `financial_search(query, top_k?)`
- `read_news_article(article_id)`
- `financial_calculate(**kwargs)`

## OUTPUT FORMAT

Once you have read enough to satisfy the constraints above, write a short evidence block — **one line per required field** — in this exact format, BEFORE calling `financial_calculate`:

```
<field_name>: <specific evidence found, with a number where possible> -> <chosen value>
```

Each line's evidence must be specific to that field's own segment/driver (unit sales, pricing, regional trends, cost data, management guidance) — it must NOT be a residual you derived by picking a total revenue or EPS figure first and back-solving this field to match it. If you genuinely found no segment-specific evidence for a field, say so explicitly and state your fallback (e.g. "no segment-specific news found; carrying forward Q-4 trend") rather than silently reverse-engineering a number from a total.

Then formulate exact numerical estimates for the required fields, consistent with the evidence block above:
- `_yoy` fields must be decimals (e.g., `0.05` for +5%).
- `_bps` fields must be integers (e.g., `-20` for a 20 basis point change).

Call `financial_calculate` with these exact estimates. As soon as it returns successfully, immediately output the final structured JSON below — do not output any further analytical text, and do not call any further tools. Assemble it as follows:
- `fundamental_predictions` = the exact arguments you passed to `financial_calculate`
- `calculated_results_summary` = the `income_statement` field returned by the tool

Do NOT output any other text or explanations outside of this JSON.

```json
{
  "fundamental_predictions": {
    "<key_1>": <value>,
    "<key_2>": <value>
  },
  "calculated_results_summary": {
    "key_metrics": {
      "eps": <value_from_tool>,
      "total_revenue": <value_from_tool>
    },
    "detail": {
      "products_revenue": <value_from_tool>,
      "gross_profit": <value_from_tool>,
      "...": "..."
    }
  }
}
```
\end{lstlisting}

\begin{lstlisting}[style=prompt, caption={The user prompt for \texttt{MSFT}, fiscal 2026Q3.}, label={lst:user_prompt}]
Ticker: MSFT
Forecast target fiscal quarter: 2026Q3
News window: 2026-01-28 to 2026-04-28. financial_search will only return articles with pub_date in this range.
Required reading:
  - Earnings call transcript (Q-1): quarter=Q2_2026
  - Financial report sections (Q-1): quarter=Q2_2026
  - Financial report sections (Q-4): quarter=Q3_2025
You MAY also read the intervening quarters (Q-2=Q1_2026, Q-3=Q4_2025) — their financial reports — to judge trend and seasonality. This matters here: you must forecast the WHOLE income statement yourself — every operating and non-operating line down to pretax income, PLUS the below-pretax lines (GAAP effective tax rate and diluted share count). Use the last four quarters to judge each line's trend/seasonality, and the GAAP tax/share history shown with the prediction fields to anchor the tax rate and share count.

All YoY values are relative to the same fiscal quarter one year ago (Q-4 = year-ago baseline).

=== Q-4 BASELINE (from most recent 10-Q filing) ===
  product_revenue: 15,319.0000
  service_and_other_revenue: 54,747.0000
  product_cogs: 3,037.0000
  service_and_other_cogs: 18,882.0000
  rd: 8,198.0000
  sales_and_marketing: 6,212.0000
  general_and_administrative: 1,737.0000
  other_expense_net: 623.0000
  income_taxes: 5,553.0000
  shares: 7,461.0000

Read the required filings above first (ECT and financial report sections), then use financial_search to discover relevant recent news. When ready, call financial_calculate with your predicted driver values to see the full computed income statement.

PREDICTION FIELDS — you must predict EXACTLY these fields, no more and no fewer.
Only these fields are valid inputs to financial_calculate; do not invent or add others:
  product_revenue_yoy: <float>  // Product revenue YoY growth rate vs Q-4. Decimal fraction. e.g. 0.05 = +5% YoY, -0.03 = -3% YoY, 0.0 = flat.
  service_and_other_revenue_yoy: <float>  // Service and other revenue YoY growth rate vs Q-4. Decimal fraction. e.g. 0.05 = +5% YoY, -0.03 = -3% YoY, 0.0 = flat.
  cogs_margin_delta_bps: <float>  // Combined cost of revenue as % of total revenue, delta vs Q-4. Basis points. e.g. -20 = margin improved 20 bps, 50 = cost ratio rose 50 bps, 0 = unchanged.
  rd_yoy: <float>  // R&D expense YoY growth rate vs Q-4. Decimal fraction. e.g. 0.05 = +5% YoY, -0.03 = -3% YoY, 0.0 = flat.
  sales_and_marketing_yoy: <float>  // Sales and marketing expense YoY growth rate vs Q-4. Decimal fraction. e.g. 0.05 = +5% YoY, -0.03 = -3% YoY, 0.0 = flat.
  general_and_administrative_yoy: <float>  // General and administrative expense YoY growth rate vs Q-4. Decimal fraction. e.g. 0.05 = +5% YoY, -0.03 = -3% YoY, 0.0 = flat.
  other_expense_net_abs: <float>  // other_expense_net absolute dollar value for the target quarter (predict the amount directly; 0 if none expected). Dollar amount for the target quarter, SAME units/scale as the Q-4 baseline above (not a growth rate). Predict the value directly; use 0 if none is expected. Sign matters (a loss/expense may be negative).
  income_taxes_abs: <float>  // income_taxes absolute dollar value for the target quarter (below-pretax line; predict directly, 0 if none expected).. Dollar amount for the target quarter, SAME units/scale as the Q-4 baseline above (not a growth rate). Predict the value directly; use 0 if none is expected. Sign matters (a loss/expense may be negative).
  shares_abs: <float>  // shares: diluted weighted-average share count for the target quarter (predict the count directly; drifts slowly with buybacks/issuance).. Dollar amount for the target quarter, SAME units/scale as the Q-4 baseline above (not a growth rate). Predict the value directly; use 0 if none is expected. Sign matters (a loss/expense may be negative).  [GAAP history — last_q=7460, ttm=7462, q4=7461]

=== UNCERTAINTY (additional required output) ===
After financial_calculate returns, you must ALSO state an 80% prediction interval for the GAAP diluted EPS it computed. Add ONE extra key to the final JSON, alongside the keys the system prompt specifies:

  "eps_interval": { "p10": <float>, "p90": <float> }

Semantics: there should be a 10% chance the actual reported GAAP diluted EPS comes in BELOW p10, and a 10% chance it comes in ABOVE p90.
The point forecast from financial_calculate stays exactly as computed — do NOT revise it to sit at the interval's midpoint, and do not call the calculator again.
Both under- and over-confidence are penalised: the interval is scored on how often the actual lands inside it AND on its width, so a deliberately wide interval scores no better than a well-judged one.
\end{lstlisting}

\stopcontents[sections]
\end{document}